\documentclass[a4paper,conference]{IEEEtran}

\usepackage{cite}
\usepackage{amsmath,amssymb,amsfonts}
\usepackage{algorithmic}
\usepackage{graphicx}
\usepackage{textcomp}
\usepackage{xcolor}

\usepackage{subfigure,url,float,bm,multirow,booktabs}

\usepackage{tabularx}

\IEEEoverridecommandlockouts
\usepackage{tikz}
\usepackage{lipsum}

\begin{document}

\title{RONELD: Robust Neural Network Output Enhancement for Active Lane Detection}

\author{\IEEEauthorblockN{Zhe Ming Chng}
\IEEEauthorblockA{\textit{Georgia Institute of Technology} \\
Atlanta, USA\\
zchng3@gatech.edu}
\and
\IEEEauthorblockN{Joseph Mun Hung Lew}
\IEEEauthorblockA{\textit{Aviation A.I. Lab Pte. Ltd.} \\
Singapore, Singapore\\
joseph.lew@aviation-ai-lab.com}
\and
\IEEEauthorblockN{Jimmy Addison Lee}
\IEEEauthorblockA{\textit{Aviation A.I. Lab Pte. Ltd.} \\
Singapore, Singapore\\
jimmy.lee@aviation-ai-lab.com}}

\maketitle
\begin{abstract}
Accurate lane detection is critical for navigation in autonomous vehicles, particularly the active lane which demarcates the single road space that the vehicle is currently traveling on. Recent state-of-the-art lane detection algorithms utilize convolutional neural networks (CNNs) to train deep learning models on popular benchmarks such as TuSimple and CULane. While each of these models works particularly well on train and test inputs obtained from the same dataset, the performance drops significantly on unseen datasets of different environments. In this paper, we present a real-time robust neural network output enhancement for active lane detection (RONELD) method to identify, track, and optimize active lanes from deep learning probability map outputs. We first adaptively extract lane points from the probability map outputs, followed by detecting curved and straight lanes before using weighted least squares linear regression on straight lanes to fix broken lane edges resulting from fragmentation of edge maps in real images. Lastly, we hypothesize true active lanes through tracking preceding frames. Experimental results demonstrate an up to two-fold increase in accuracy using RONELD on cross-dataset validation tests.

\end{abstract}

\begin{IEEEkeywords}
lane detection, lane tracking, autonomous driving
\end{IEEEkeywords}

\section{Introduction}
\label{sec:intro}

In recent years, autonomous driving has received much attention in computer vision and robotics research, at both academic and industrial levels. A critical step in autonomous driving is the recognition of the operating environment by the vehicle. Road lane markings form an integral component of this operating environment. In particular, active lane markings serve as significant cues for constraining the maneuver of vehicles on roads by indicating the active lane, which is the single usable road space by the vehicle, that serves as input for lateral steering control to avoid collisions with other road users. Despite the pressing need for accurate and reliable lane detection to enable successful autonomous vehicles, detecting lanes has remained challenging throughout the years. One reason is the rather simple and homogeneous appearance of lane markings which lacks distinctive features. Other obstacles, such as weather and illumination conditions, also plague lane detection research. Furthermore, lane detection scenarios occur in diverse driving environments, various road surface conditions, and in real-time, which necessitates a robust and low computational cost algorithm for successful lane detection on autonomous vehicles.

To address the lane detection problem, deep learning models have gained popularity in recent lane detection literature~\cite{Garnett2019, Ghafoorian2018, Gansbeke2019,Hou2020}. Contemporary lane detection algorithms based on end-to-end deep learning models have shown great promise in addressing the lane detection problem~\cite{Pan2018, Hou2019, Garnett2019}, achieving competitive results against traditional lane detection methods and are more robust to a greater range of driving conditions. However, it is observed that many of these models still do not perform well on datasets that differ significantly from their train sets, \textit{e.g.} varying road surface conditions and different road lane markings. False positives, otherwise known as noise, undetected lanes, and broken lane edges are common which lead to accuracy degradation and instability in the control of autonomous vehicles in these situations.

\begin{figure}[t]
	\centering
	\subfigure[Before]
	{
		\includegraphics[width=38mm]{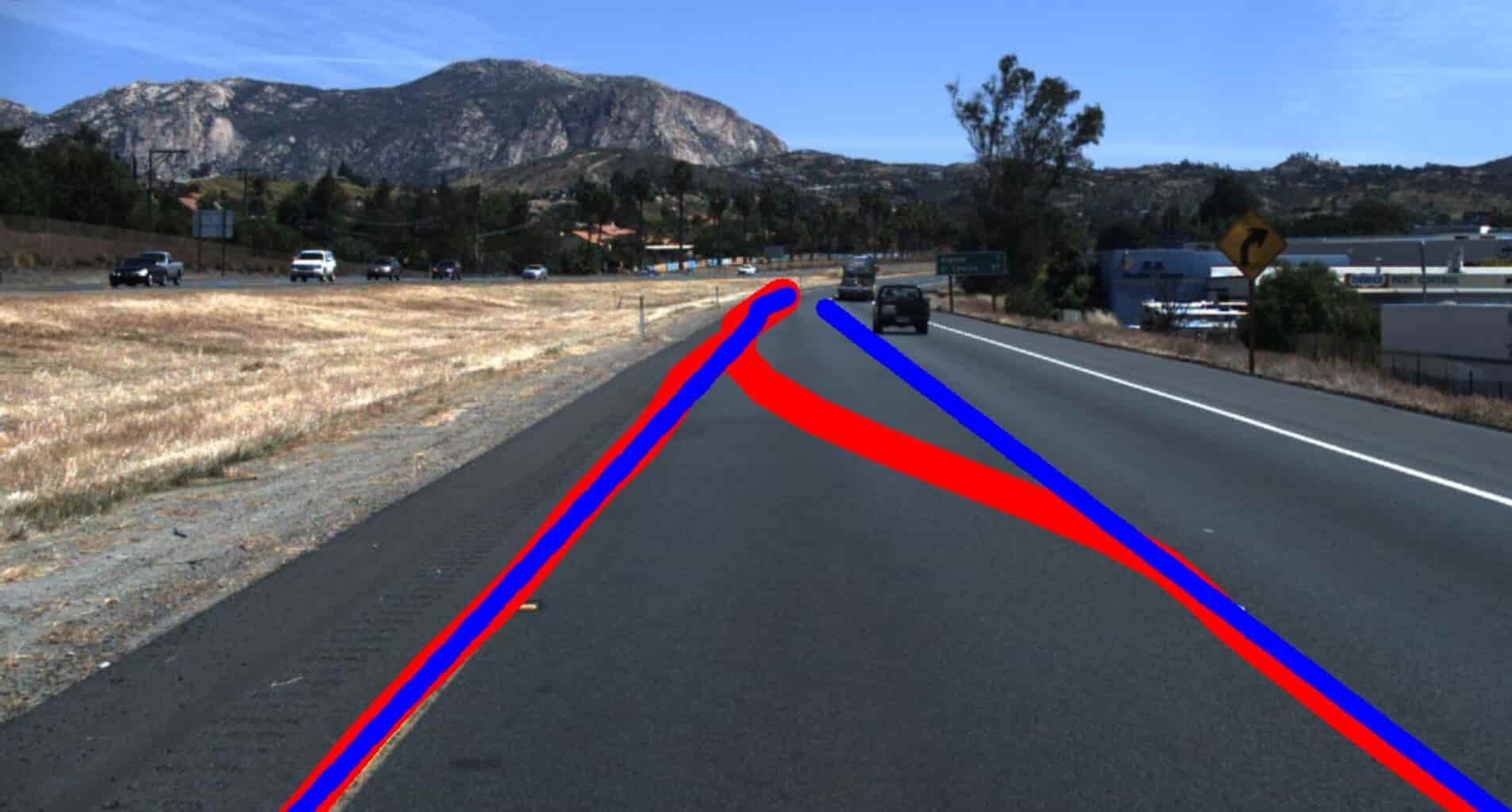}\label{fig1a}
	}
	\subfigure[After]
	{
		\includegraphics[width=38mm]{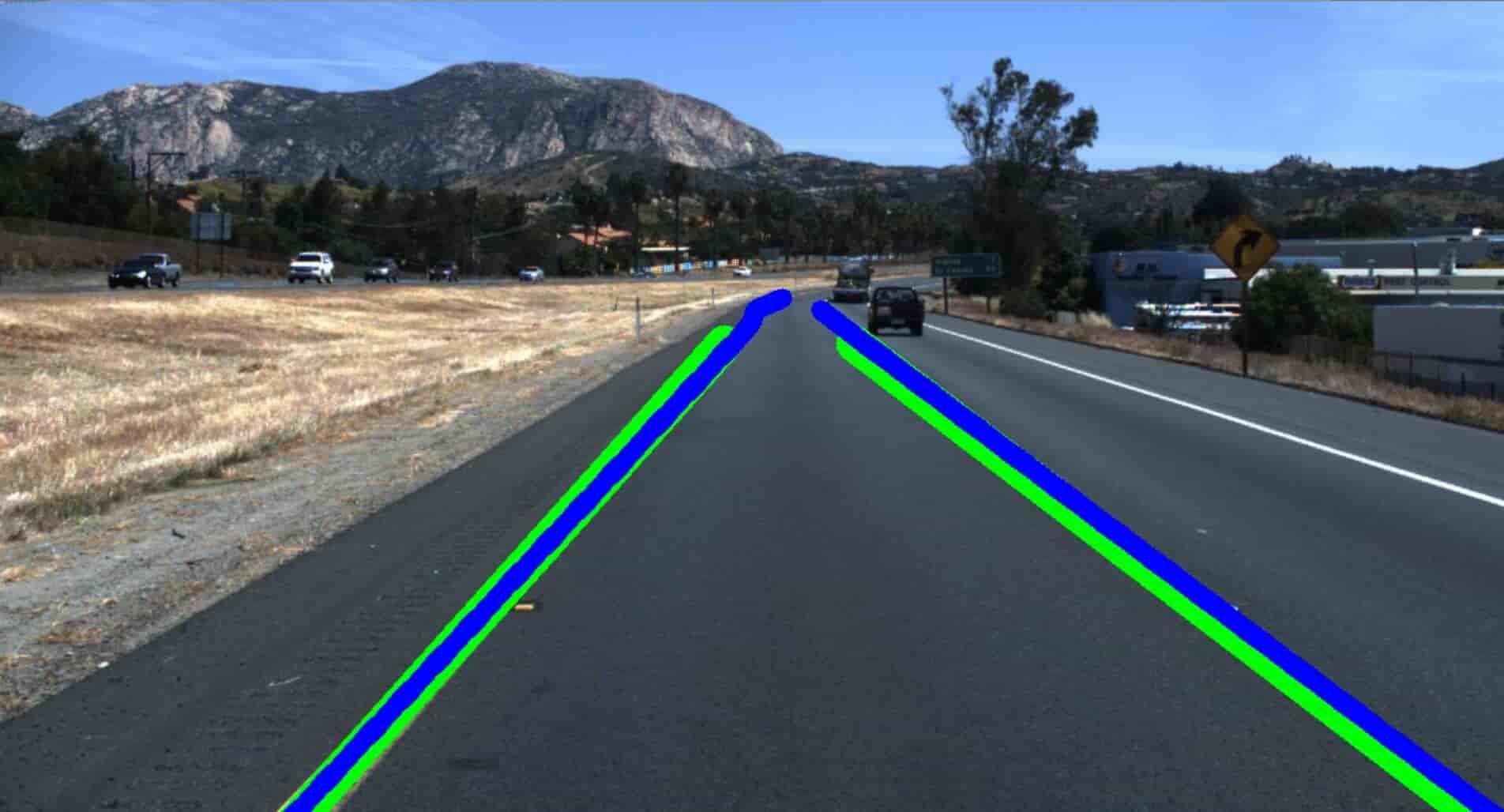}\label{fig1b}
	}
	\subfigure[Before]
	{
		\includegraphics[width=38mm]{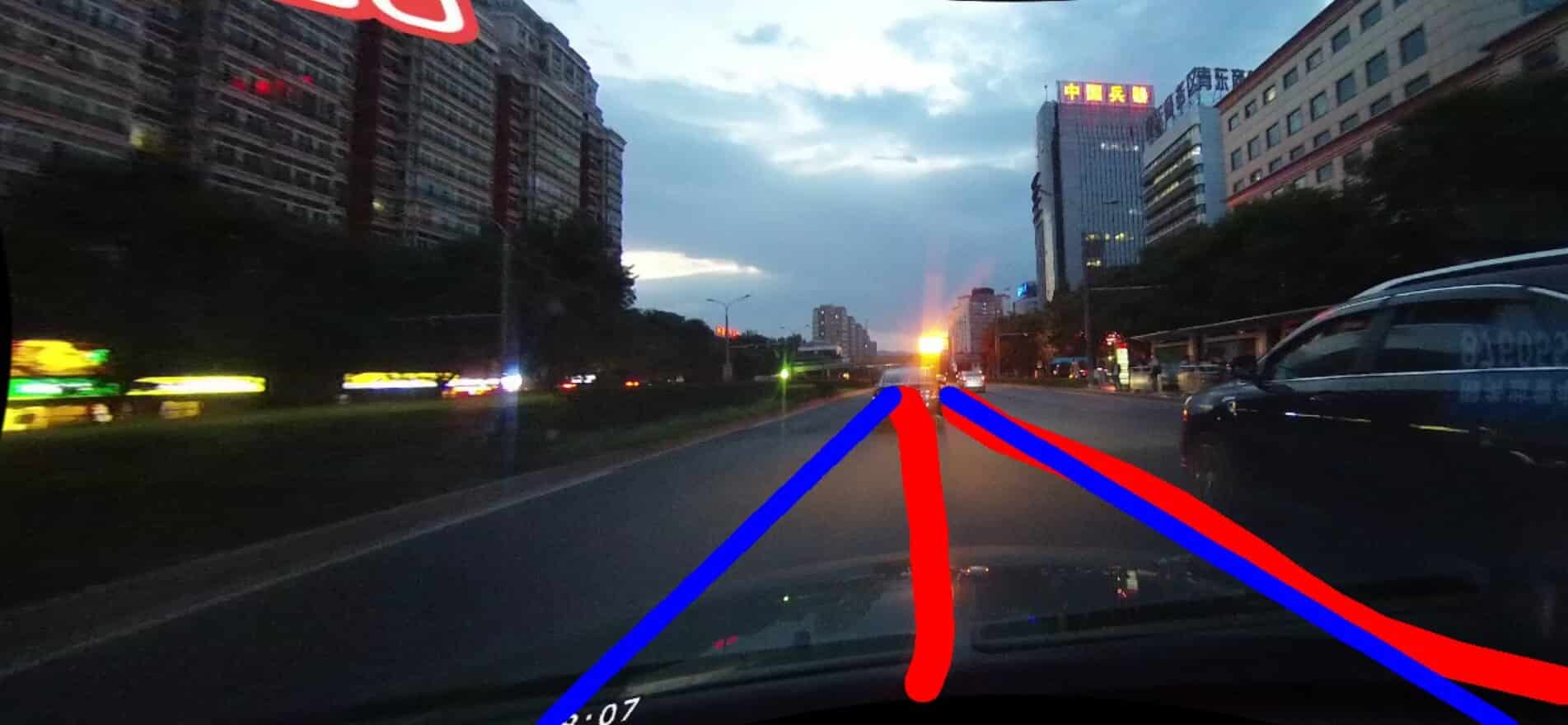}\label{fig1c}
	}
	\subfigure[After]
	{
		\includegraphics[width=38mm]{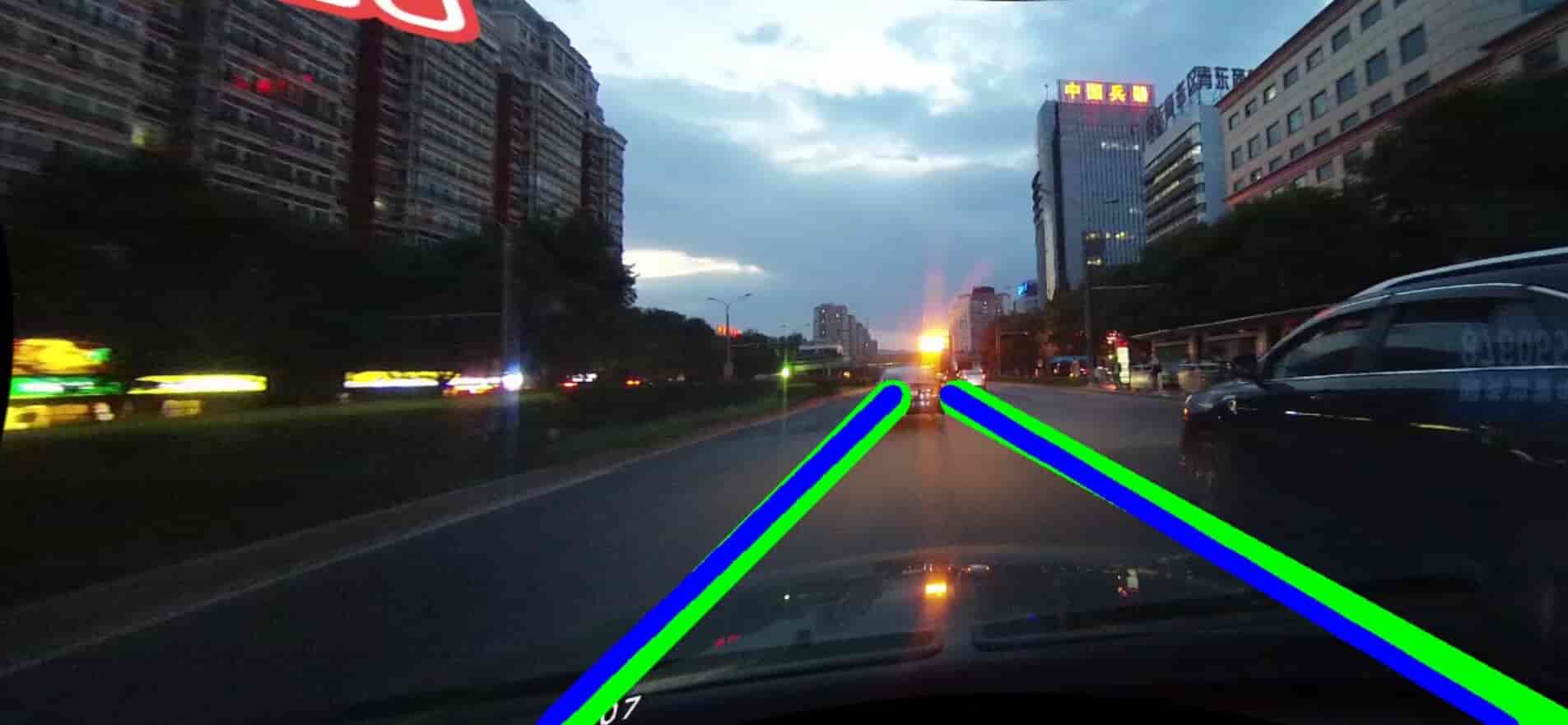}\label{fig1d}
	}
	\caption{Two SCNN lane detection results on the TuSimple~\cite{TuSimple-2019} and CULane~\cite{Pan2018} test sets are shown in (a) and (c) respectively, highlighted in red, while ground truths are highlighted in blue. Corresponding results after RONELD are shown in (b) and (d) respectively, highlighted in green.}\label{fig1}\end{figure}

In this paper, we aim to address this issue by proposing a robust neural network output enhancement for active lane detection (RONELD) method to strive for a robust, low computational cost and real-time solution for use together with deep learning models on autonomous vehicles. It is motivated by the poor performance of existing deep learning models on new, unseen datasets that makes them problematic to use on autonomous vehicles which rely heavily on accurate lane detection. Our method is built on the observation that accuracy performance can be improved through enhancement of the predicted lane markings from existing deep learning model probability map outputs. In particular, the accuracy performance can be significantly increased on datasets which differ greatly from the train set of the deep learning model. RONELD is intended as a turnkey solution leveraging probability map outputs from deep learning models to optimize active lane detection for more stable and robust active lanes that are better suited for autonomous driving applications. In addition, it is a low computational time solution, making it suitable for real-time use on autonomous vehicles. To verify the usefulness of RONELD, we test it on two state-of-the-art deep learning models, Spatial CNN (SCNN)~\cite{Pan2018} and ENet-SAD~\cite{Hou2019}, and record the resulting accuracy and processing time of RONELD. Our experiments successfully demonstrate the fast runtime and effectiveness of using RONELD on the two popular state-of-the-art deep learning models through the increased accuracy performance. In Fig.~\ref{fig1}, we show two simple before and after results of applying RONELD on the SCNN deep learning model probability map output.

The rest of the paper is organized as follows. Section~\ref{sec:relatedworks} discusses related work. In Section~\ref{sec:methodology}, we explain our methodology in four parts: Adaptive lane point extraction, curved lane detection, lane construction and tracking preceding frames. Section~\ref{sec:experiment} compares experimental results with benchmarks, and Section~\ref{sec:conclusion} concludes our work.

\section{Related work}\label{sec:relatedworks}
\textbf{Traditional lane detection.} Traditional lane detection methods~\cite{Wu-2014,Tan2014,Kaur2015,Deusch2012} rely on hand-crafted features such as color-based features~\cite{Chiu2005}, bar filter~\cite{Teng2010}, ridge features~\cite{Lopez2010}, hough transform~\cite{Liu2010,Zhou2010}, random sample consensus (RANSAC)~\cite{Borkar2009, Aly2008}, \textit{etc.}, to identify lane segments. Tracking techniques such as particle or Kalman filters~\cite{Teng2010,Danescu2009,Borkar2009} are used as a final stage for lane tracking to map the lanes onto the current frame. Loose \textit{et al.} combined the Kalman and particle filters into a Kalman Particle Filter~\cite{Loose2009} for lane detection on non-marked rural roads. In general, most of these traditional methods based on hand-crafted features lack robustness and can only solve the lane detection problem in limited scenarios or require strict lane assumptions, \textit{e.g.} lanes are straight~\cite{Li-2015,Niu2016} and parallel~\cite{Jiang2010,Nieto2008}. These conditions are not always valid, particularly in complicated urban driving environments or scenes with poor weather and road conditions where issues such as varying road surface conditions (\textit{e.g.} faded lane markings, discolored road surfaces), different lane marking colors, and visibility significantly impact the accuracy of traditional lane detection methods.

\textbf{Deep learning based lane detection.} After demonstrating compelling results in many other computer vision problems~\cite{Zou2020}, deep learning methods have been introduced to replace traditional hand-crafted feature-based lane detection algorithms in addressing the lane detection problem~\cite{Lee2017, Garnett2019, Ghafoorian2018, Gurghian2016}. One common approach is to treat lane detection as a semantic segmentation task and use end-to-end deep learning models to formulate dense predictions, \textit{i.e.} predict a label for each pixel in the image to indicate if it is part of a lane marking~\cite{Pan2018, Hou2019, He-2016}. There have also been some methods introduced that use an instance segmentation approach as well, \textit{i.e.} treat each lane as its own instance~\cite{Chang2019, Nevan2018}. He \textit{et al.} introduced a Dual-View CNN (DVCNN)~\cite{He-2016} method which uses front and top view images simultaneously to eliminate false positives and remove non-club-shaped structures respectively. Lee \textit{et al.} proposed a vanishing point guided network (VPGNet)~\cite{Lee2017} to address lane detection under low illumination conditions by detecting lane and road markings as well as the vanishing point in a multi-task network. Later on, Pan \textit{et al.} proposed SCNN~\cite{Pan2018}, which generalized deep layer-by-layer convolutions to slice-by-slice convolutions within feature maps, thus enabling message passing between pixels across rows and columns in a layer. It is designed for long continuous structured or large objects, with strong spatial relationships but less appearance clues (\textit{e.g.} traffic lanes). Recently, a self attention distillation (SAD) method~\cite{Hou2019} was proposed, incorporated with the lightweight ENet~\cite{Paszke2016}, ResNet-18~\cite{resnet}, and ResNet-34~\cite{resnet} models. In particular, the SAD-incorporated ENet model, titled ENet-SAD, runs 10 times faster than SCNN while achieving comparable performance in popular benchmarks such as CULane~\cite{Pan2018} and TuSimple~\cite{TuSimple-2019}.

\begin{figure}
    \centering
    \subfigure[Input]{
        \hspace{-2.5mm}
        \includegraphics[width=28mm,height=15.75mm]{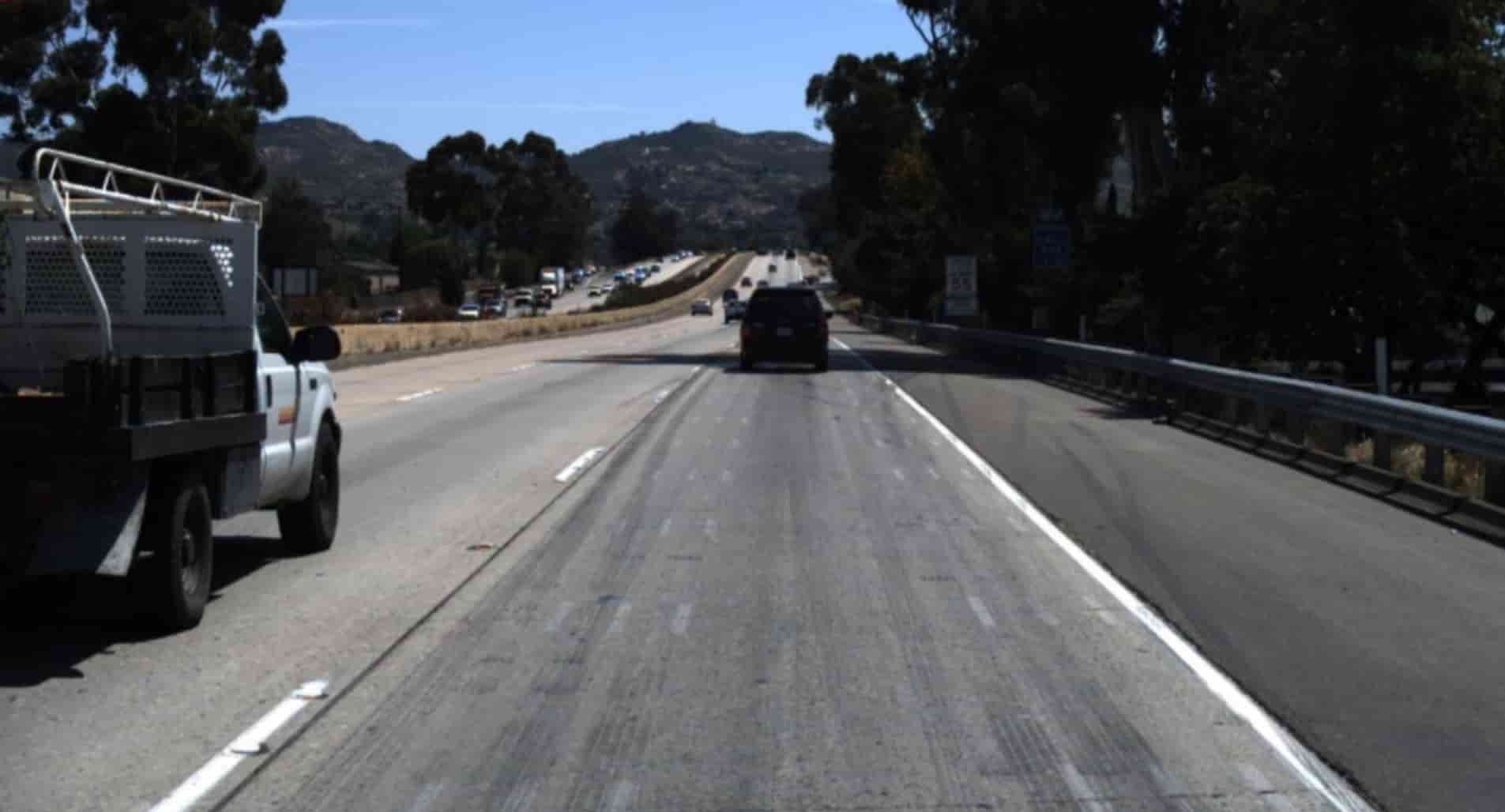}
        \hspace{-2.5mm}
    }
    \subfigure[SCNN]{
        \hspace{-2.5mm}
        \includegraphics[width=28mm,height=16mm]{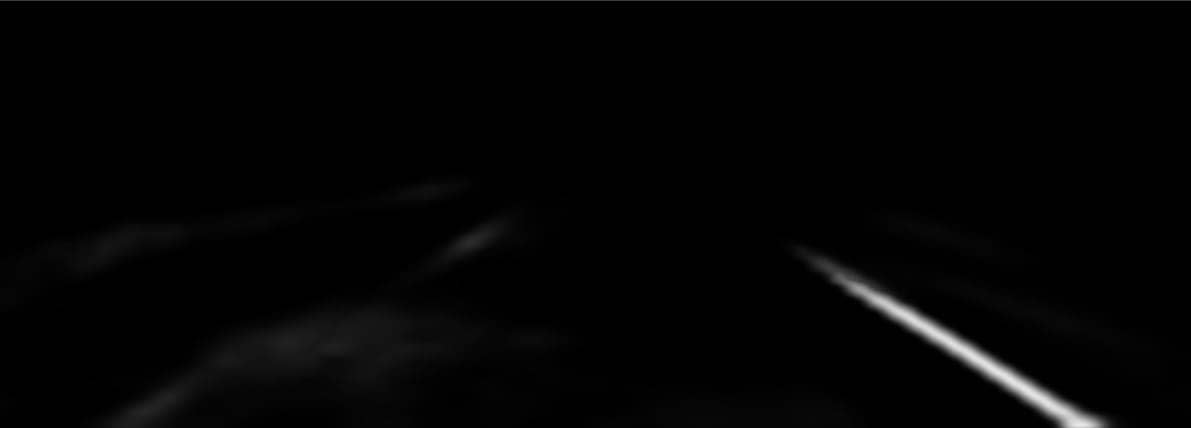}
        \hspace{-2.5mm}
    }
    \subfigure[ENet-SAD]{
        \hspace{-2.5mm}
        \includegraphics[width=28mm,height=16mm]{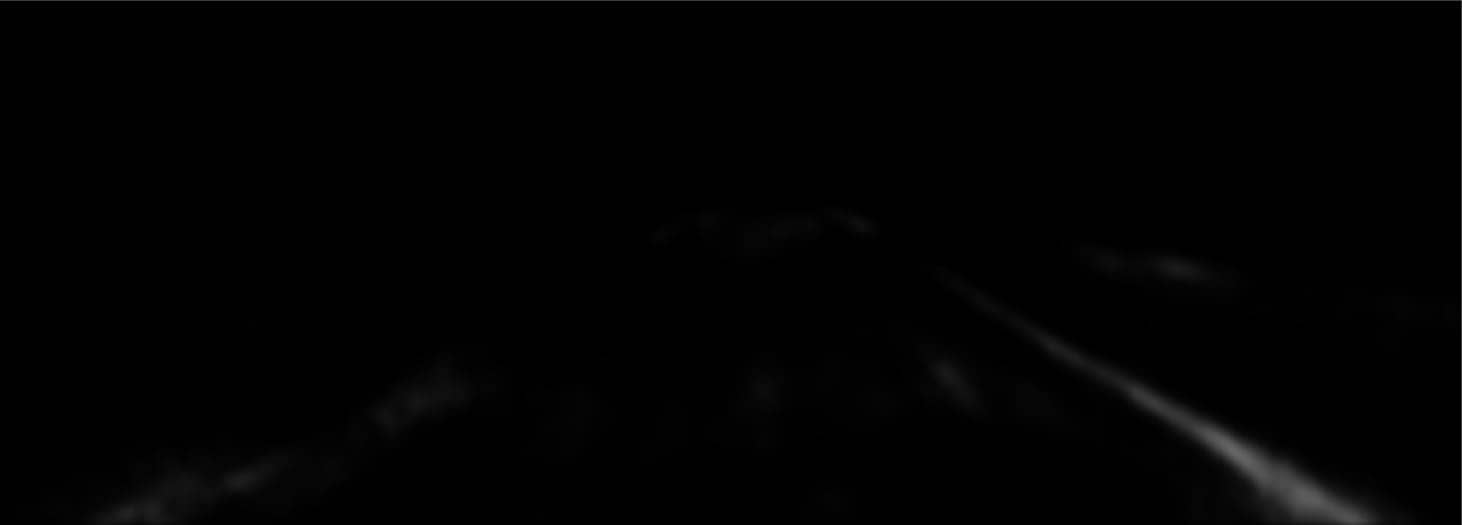}
        \hspace{-2.5mm}
    }
    \subfigure[Input]{
        \hspace{-2.5mm}
        \includegraphics[width=28mm,height=16mm]{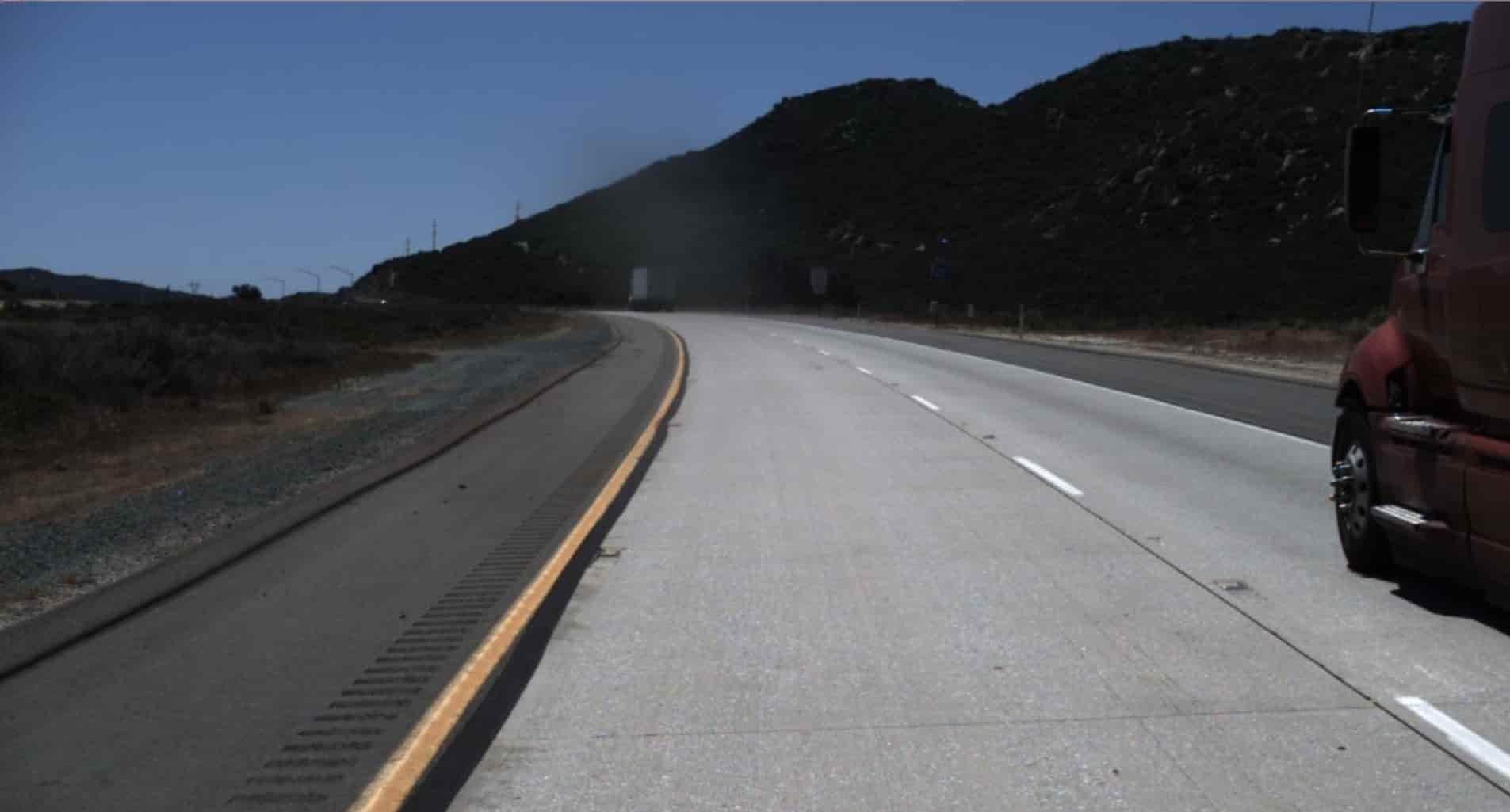}
        \hspace{-2.5mm}
    }
    \subfigure[SCNN]{
        \hspace{-2.5mm}
        \includegraphics[width=28mm,height=16mm]{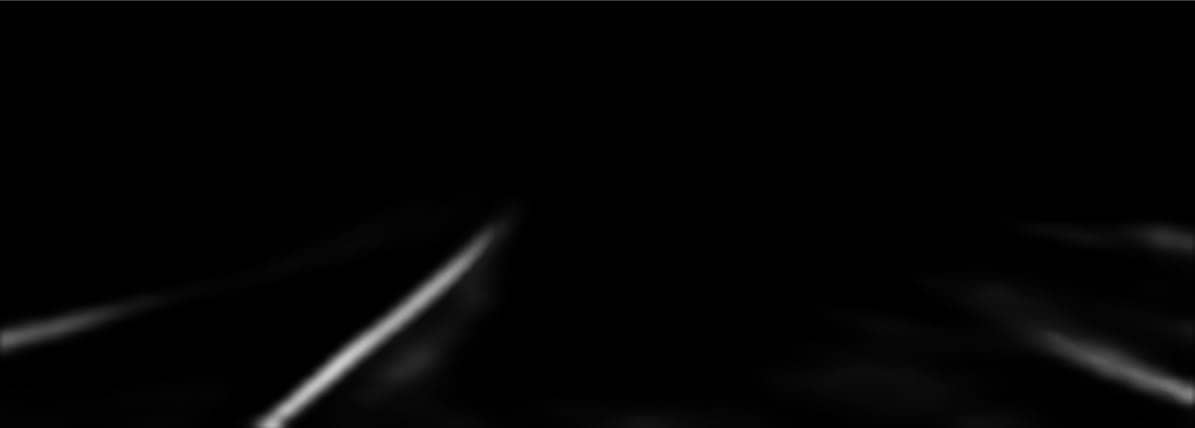}
        \hspace{-2.5mm}
    }
    \subfigure[ENet-SAD]{
        \hspace{-2.5mm}
        \includegraphics[width=28mm,height=16mm]{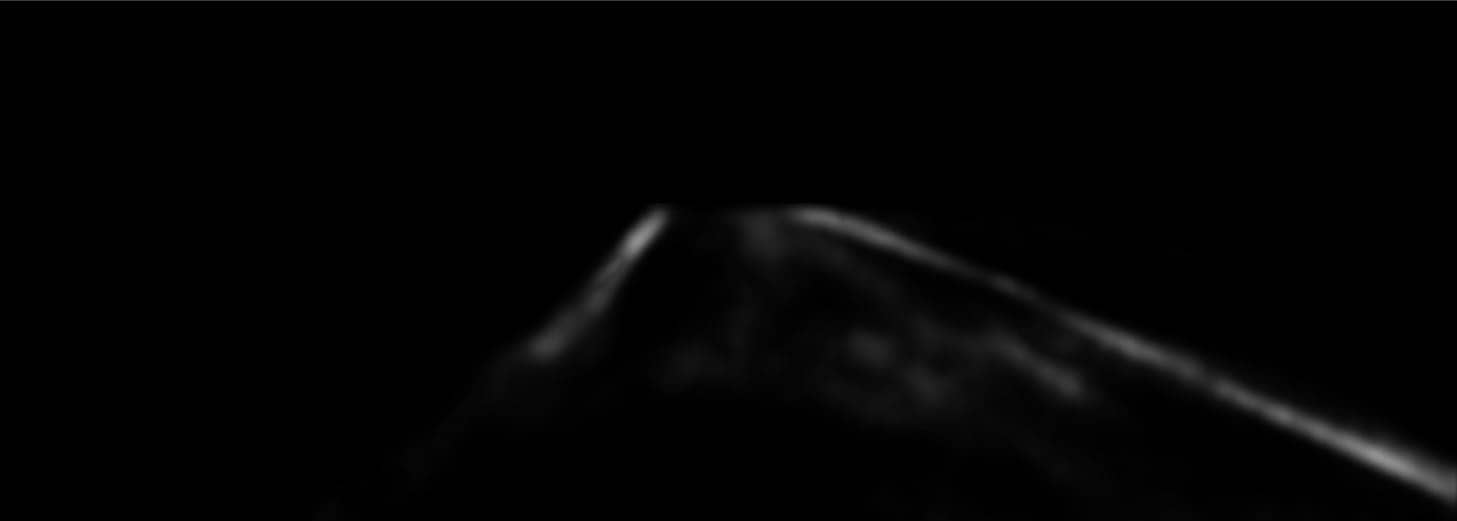}
        \hspace{-2.5mm}
    }
    \caption{Probability map outputs from the CULane-trained SCNN and ENet-SAD deep learning models on sample images from the TuSimple test set. }\label{proboutputfig}
\end{figure}

Although the aforementioned deep learning methods provide promising lane detection results on trained datasets, their inflexibility presents challenges when road conditions deviate from their train sets. This is also a concern when the lane markings are obscured or degraded due to fading, stains, shadows, or occlusions. These issues makes it difficult for these models to be applied on autonomous vehicles that might encounter new, unseen environments. In Fig.~\ref{proboutputfig}, we include probability map outputs from the CULane-trained SCNN and ENet-SAD models on unseen TuSimple test set images to illustrate this issue. Alternatively, the deep learning models require extensive train sets to account for the myriad of different possible environments that autonomous vehicles might encounter which is expensive and time-consuming. 

To tackle this problem, some methods using various techniques to enhance lane detection outputs from deep learning models were proposed~\cite{Ko2020,Kim2014,Lee2017}, usually exploiting some geometric properties (\textit{e.g.} vanishing points). However, these methods are usually paired for use with specific deep learning models or lack robustness, which make them unsuitable for autonomous driving applications with other existing models.

\begin{figure*}
	\centering
	\includegraphics[width=\textwidth]{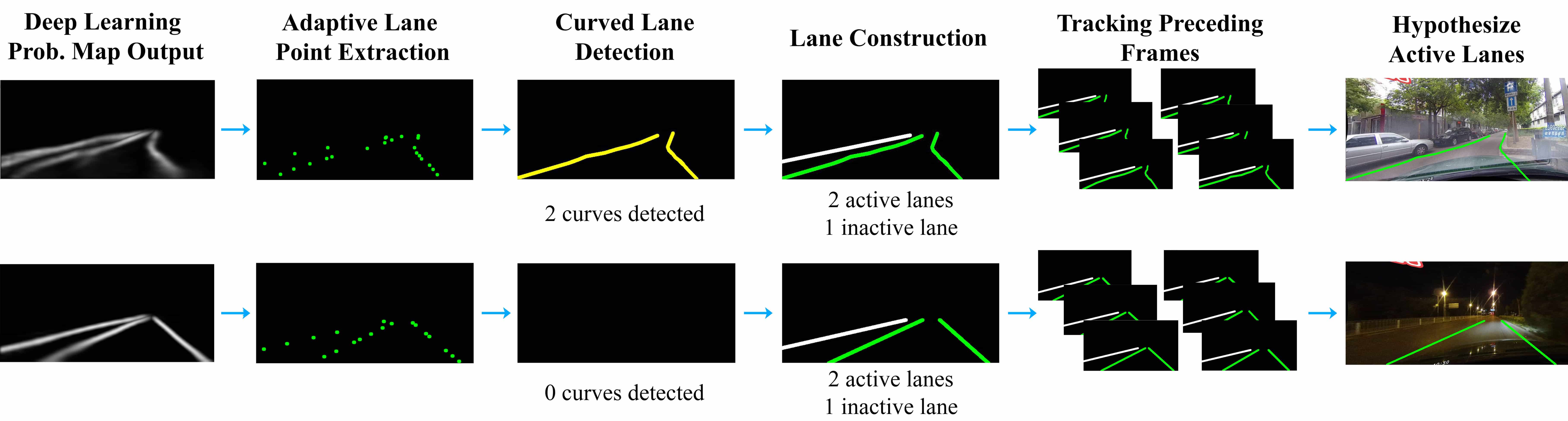}
	\caption{RONELD's process workflow to hypothesize straight and curved active lanes from probability map outputs obtained by running the SCNN deep learning model on two sample images.}\label{fig2}\end{figure*}

\section{Methodology}
\label{sec:methodology}
In this section, We discuss our RONELD method with illustration from the process workflow diagram presented in Fig.~\ref{fig2}.

\subsection{Adaptive Lane Point Extraction}
We first extract lane points from the probability map output generated from a deep learning model (\textit{e.g.} SCNN). For robustness and to exclude low-confidence noise, while being able to accurately detect lane points in the context of the current frame's probability map output, we search only salient points by picking points in excess of an adaptive confidence threshold. This confidence threshold is adapted based on the highest confidence point in the probability map outputs.

After a salient point is found on a detected lane marking, we search the subsequent $\frac{h}{20}$ rows and neighbouring $\frac{w}{2}$ columns, where $h$ is the height of the probability map output and $w$ is the width of the probability map output. We search only neighbouring points to exclude extraneous objects and noise in other parts of the output as well as to reduce processing times by focusing our search area around the detected salient point. We take the highest confidence point found on the lane within the search area as a lane point for the detected lane marking. This is repeated throughout the probability map outputs to identify lane points for each lane marking.

\subsection{Curved Lane Detection}
We separate detected lane markings into two broad categories: straight lanes and curved lanes. We do this to adjust for lanes that do not follow a linear model while being able to use a linear model to fix broken (undetected) lane edges in the probability map output for straight lanes. For straight lanes, we require a minimum of $n=3$ lane points to reduce the impact of noise. For curved lanes, due to their greater complexity, we define the minimum as $3n$. To differentiate between the two categories, we use the coefficient of determination $r^2$ to assess how well the points fit a linear model. The coefficient of determination measures the proportion of variance in one variable that can be explained by a linear regression model and predictor variable(s). It is calculated as follows:
\vspace{2mm}
\begin{equation}
	r^2 = \frac{[\mathrm{Cov}(X, Y)] ^ 2}{\mathrm{Var}(X)\mathrm{Var}(Y)},
\end{equation}
where $(X, Y)$ are the random variables denoting the $x$- and $y$-components of detected lane points, $\mathrm{Cov}(X,Y)$ is the covariance of X and Y and $\mathrm{Var}(X)$ is the variance of X. We compare the $r^2$ between the whole lane marking and the lane marking without the top $n$ points. If the whole lane marking has a lower $r^2$ than the truncated section, it suggests that the lane marking proves a worse fit for a linear model than this truncated section. This, in turn, implies that the top $n$ lane points are straying further from the regression line, which is characteristic of a curve. Hence, we mark this lane as a curved lane. Conversely, we mark it as a straight lane if it does not fulfill this criteria. To reduce false curve predictions in our detected lane markings, we corroborate curves in the current frame with curves in previous frames.

\subsection{Lane Construction}\label{lanesubsection}
A predicted curved lane will have its points connected by quadratic splines to form the final lane marking output. For straight lanes, we attempt to fix broken lane edges in the probability map output and remove outliers by considering the detected lane points based on a linear model with the form:
\begin{equation}
    \bm{y} = \bm{X}\bm{\beta},
\end{equation}
where $\bm{y}\in\mathbb{R}^{m \times 1}$, $\bm{X}\in\mathbb{R}^{m \times 2}$, $\bm{\beta}\in\mathbb{R}^{2 \times 1}$:
\begin{equation*}
    \bm{y} = \begin{pmatrix}
        y_1\\
        y_2\\
        \vdots\\
        y_m
    \end{pmatrix}, \,
    \bm{X} = \begin{pmatrix}
        1 & x_1\\
        1 & x_2\\
        \vdots & \vdots\\
        1 & x_m
    \end{pmatrix}, \, 
    \bm{\beta} = \begin{pmatrix}
        \beta_0\\
        \beta_1
    \end{pmatrix},
\end{equation*}
where $(x_i, y_i)$ are the $x$- and $y$-coordinate of the $i$-th detected lane point, $(\beta_0, \beta_1)$ are the $y$-intercept and gradient of the line respectively, and $m$ is the number of detected lane points.

We obtain a weighted least squared error estimate of $\bm{\beta}$, $\bm{\hat{\beta}}$, based on the sample of detected lane points by adapting weighted ordinary least squares linear regression. For our method, we set the weights as the confidence of each detected lane point in the probability map output of the deep learning model. This is to reduce the problem of heteroskedasticity, as the variance of $y$-coordinates is not constant across the range of $x$-coordinates for the detected lane points, however there is a constraint that the variance for each detected lane point reading is unknown and would depend on factors such as the accuracy of the deep learning model on that dataset. To address this, higher confidence points are assumed to be more accurately detected and hence have smaller errors with lower variances. Therefore, we allocate a higher weight to these points by virtue by their higher confidence on the output probability maps.
We search for the solution that minimizes the weighted squared error term $\|\bm{\epsilon}\|^2$ for our linear model, where $\bm{\epsilon}$ is the vector containing the weighted error term for each lane point observation, and is calculated as follows:
\begin{equation}
    \|\bm{\epsilon}\|^2 = \|\bm{C}^\frac{1}{2}(\bm{y}-\bm{X}\bm{\hat{\beta}})\|^2,
\end{equation}
where $\bm{C}\in\mathbb{R}^{m \times m}$, 
\begin{equation*}
    \bm{C} = \begin{pmatrix}
    c_1 & 0 & \dots & 0\\
    0 & c_2 & \dots & 0\\
    \vdots & \vdots & \ddots & \vdots\\
    0 & 0 & \dots & c_m
    \end{pmatrix},
\end{equation*}
and $c_i$ is the confidence of the $i$-th detected lane point. By minimizing $\|\bm{\epsilon}\|^2$ for our sample of detected lane points, we obtain $\bm{\hat{\beta}}$ as follows:
\begin{equation}
	\bm{\hat{\beta}} = \operatorname*{arg\,min}_{\bm{\beta}}||\bm{C}^\frac{1}{2}(\bm{y}-\bm{X}\bm{\beta})||^2 =  (\bm{X}^T\bm{C}\bm{X})^{-1}\bm{X}^T\bm{C}\bm{y}.
\end{equation}
From $\bm{\hat{\beta}}$, we are able to obtain the weighted least squares error gradient and $y$-intercept for the straight lane based on the extracted lane points. To reduce false positive noise from our set of straight lane points, we remove outliers, which are points that are significantly further from the regression line than other points. We do this based on the $x$-distance between each point and the regression line. This step allows us to obtain a more accurate final $\bm{\hat{\beta}}$ for the regression line which we store as the straight lane marking parameters. Using these lane parameters, we are able to fix broken lane markings due to undetected lane edges in the deep learning model probability map output.
\vspace{-0.5mm}
\subsection{Tracking Preceding Frames}
In complicated driving environments with varying weather, illumination and road conditions, the current frame may be insufficient for accurate lane detection. The lanes in the current frame may be obscured or degraded by shadows, poor road conditions (\textit{e.g.} stains, fading), or occlusions. To address this and minimize distortions arising from incorrectly identified lanes in the probability map, we track lanes in preceding frames and map them to lanes in the current frame to hypothesize stable and robust active lanes. We do this by calculating the root mean square (RMS) $x$-distance, $\zeta(L_1, L_2)$, between previous and current lane markings, shown as follows:
\begin{equation}
	\zeta(L_1,L_2) = \sqrt{\frac{1}{h}\int_{0}^{h} \smash[b]{(\underset{x_1}{\underbrace{\frac{y-b_1}{a_1}}}} - \smash[b]{\underset{x_2}{\underbrace{\frac{y-b_2}{a_2}}}}) ^ 2 dy},
	\vspace{1mm}
\end{equation}
where $L_1$, $L_2$ are the two lane markings under consideration. $x_k$, $a_k$, $b_k$ are the $x$-coordinate, gradient and $y$-intercept of $L_k$ respectively, where $k \in \{1,2\}$. For lanes with $\zeta(L_1, L_2)\leq{\frac{w}{200}}$, we consider them as the same lane. If there is more than one lane in the preceding frame that matches the current lane, we match the current lane with the previous lane that has the smallest $\zeta(L_1,L_2)$ value to ensure that each current lane marking is matched to only one previous lane marking.

We track lane markings and assign them weights based on their appearance in previous frames. We do this to map lane markings onto the current frame even if the lanes are undetected for some intermediate frames, due to reasons such as fading, shadows or occlusions, while remaining robust to changes in the driving environment. Lane markings with a greater number of high confidence points and labelled as potential active lane markings in the current frame are given a higher weight increment due to the increased likelihood of them forming the active lane. Meanwhile, for lane markings that do not appear in the current frame, we decrease their weight exponentially to remain responsive to the changing driving environment. The following equation is used to calculate the weight of lane marking $L$, $W_L$:
\begin{equation}
	W_L = \sum_{f \,\in\, \mathbf{F}}{e^{-d}\,\psi\,c^f_L\,N_L(f)},
\end{equation}
where $\psi$ is the weight increment factor which is higher for identified potential active lane markings and lower for non-active lane markings, $d$ is the number of frames in which the lane marking was missing since being detected, $c^f_L$ is the RMS confidence of lane points and $N_L(f)$ is the number of lane points in lane marking $L$ in frame $f$, and $\mathbf{F}$ is all previous and current frames. We identify the potential active lane markings based on the deep learning model output and assign a higher $\psi$ value for potential active lane markings to prioritize them while recording the inactive lane markings for subsequent frames to process. We store inactive lane markings in the current frame in addition to identified active lane markings as lane markings might be identified incorrectly as the active lane marking, \textit{e.g.} due to false positive lane markings as shown in Fig.~\ref{fig1c}, and the true active lane marking might be misclassified as an inactive lane marking, hence we store inactive lane markings present as well. As the inactive lane markings are lane markings that have been identified in the current image, keeping record of them and assigning them weights helps RONELD have a better understanding of the current lane environment and remain robust to changes in the driving environment (\textit{e.g.} due to lane changes by the vehicle).

We rank lane markings based on their $W_L$ after processing the current frame. We take one lane marking each from the left and right side of the image with the highest $W_L$, and mark them as the left and right lane marking for our active lane. Finally, we use the lane marking parameters and the camera's extrinsic parameters to plot our final lane marking output, using the aforementioned linear model and quadratic spline in subsection~\ref{lanesubsection} for straight and curved lanes respectively.
	
\begin{table*}
	\small
	\caption{Basic information of datasets (TuSimple and CULane) used in our experiments.} \label{table1}
	\centering
	\vspace{-1mm}
	\begin{tabular}{c | c c c c c c|}
		\hline
		Dataset
		& \#Total
		& \#Test 
		& Resolution
		& Road Type \\
		\hline
		TuSimple
		& 6,408
		& 2,782
		& 1280$\times$720
		& Highway \\
		CULane
		& 133,235
		& 34,680
		& 1640$\times$590
		& Urban, rural, highway \\
		\hline
	\end{tabular}
    \vspace{-1mm}
\end{table*}

\vspace{-1.5mm}
\section{EXPERIMENTAL RESULTS}
\label{sec:experiment}
\vspace{-0.5mm}
\subsection{Datasets} 
\begin{figure}[t]
	\centering
	
	\includegraphics[height=18mm, width=39.6mm]{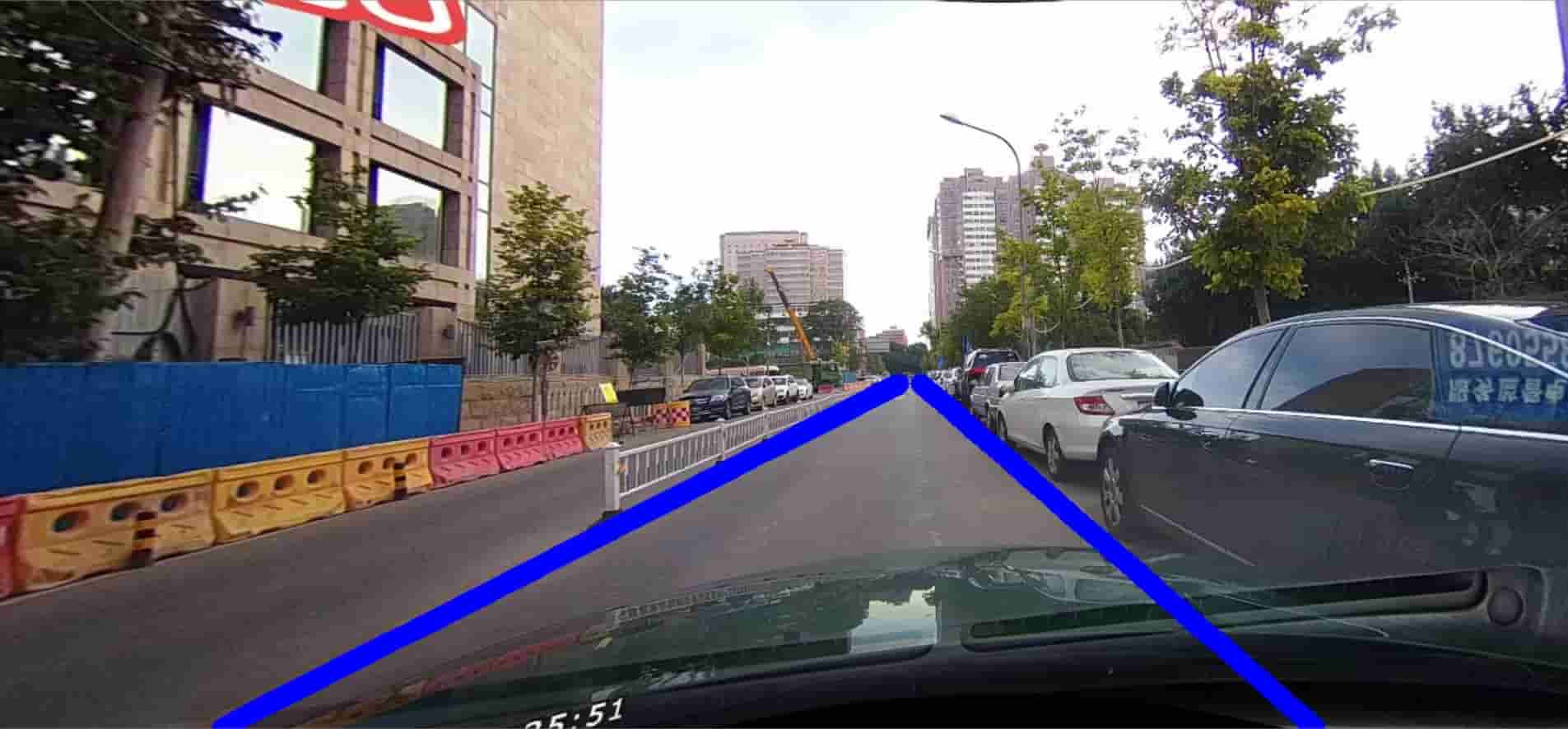}\label{culane1pic}
	\hspace{1.5mm}
	\includegraphics[height=18mm, width=32mm]{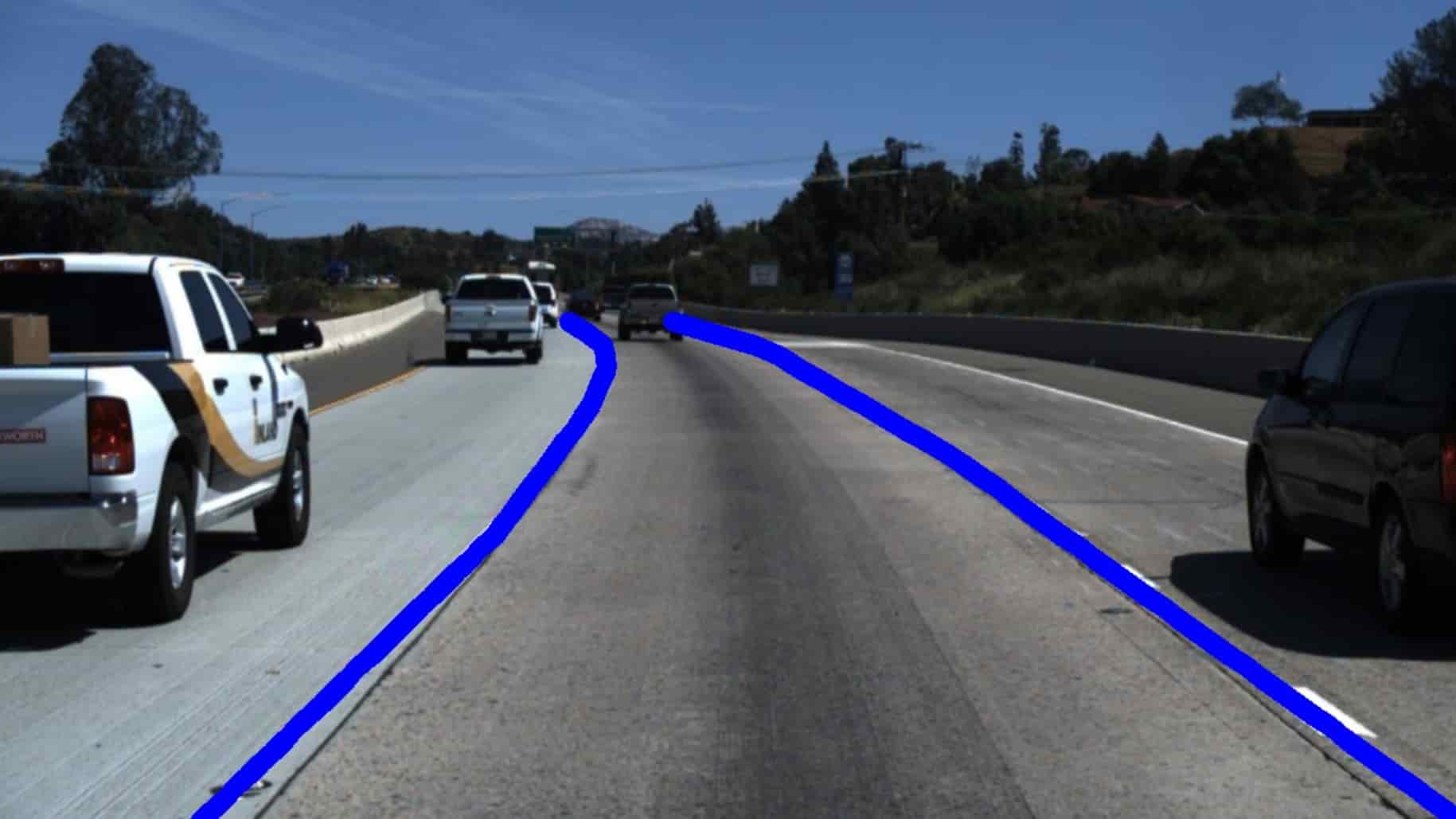}\label{tusimple1pic}
	\vspace{2mm}
	
	\includegraphics[height=18mm, width=39.6mm]{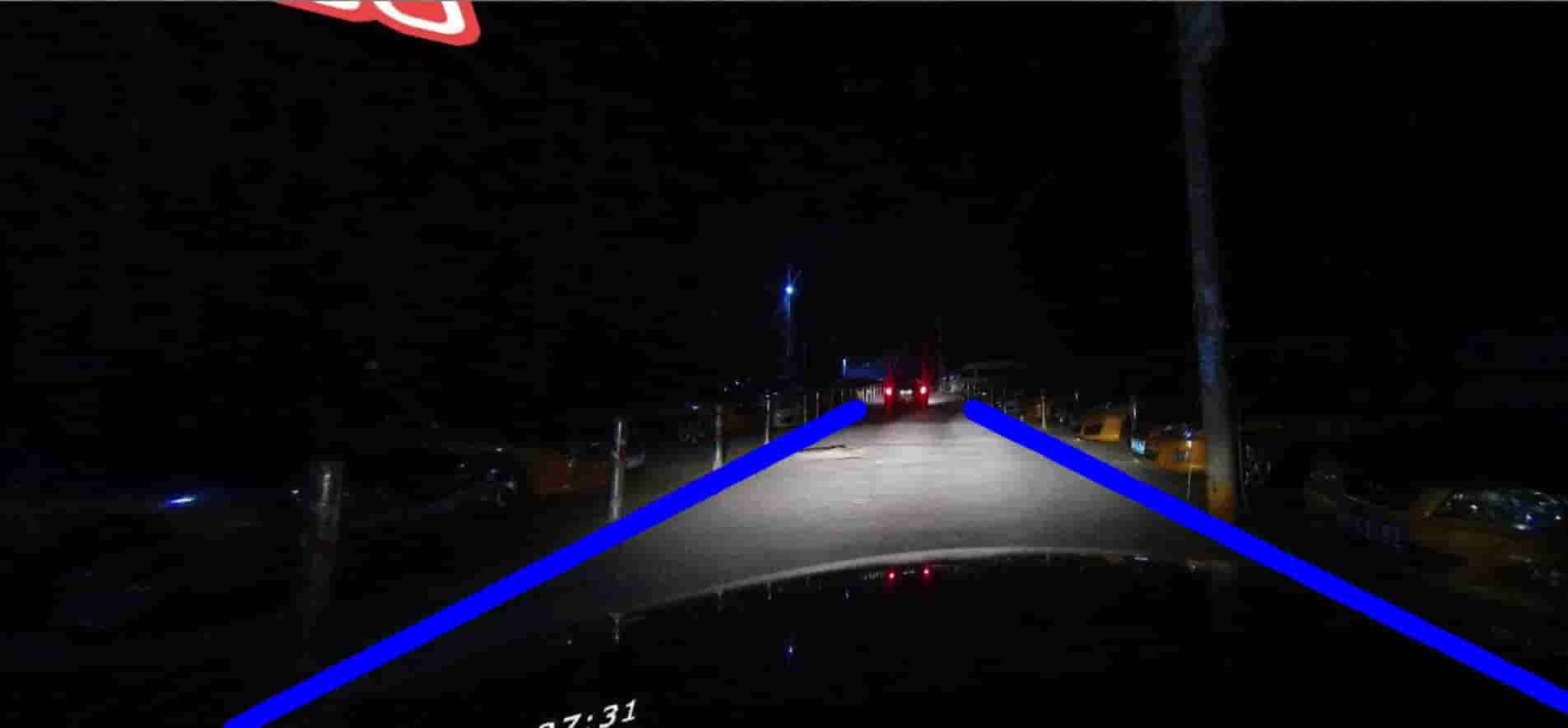}\label{culane2pic}
	\hspace{1.5mm}
	\includegraphics[height=18mm, width=32mm]{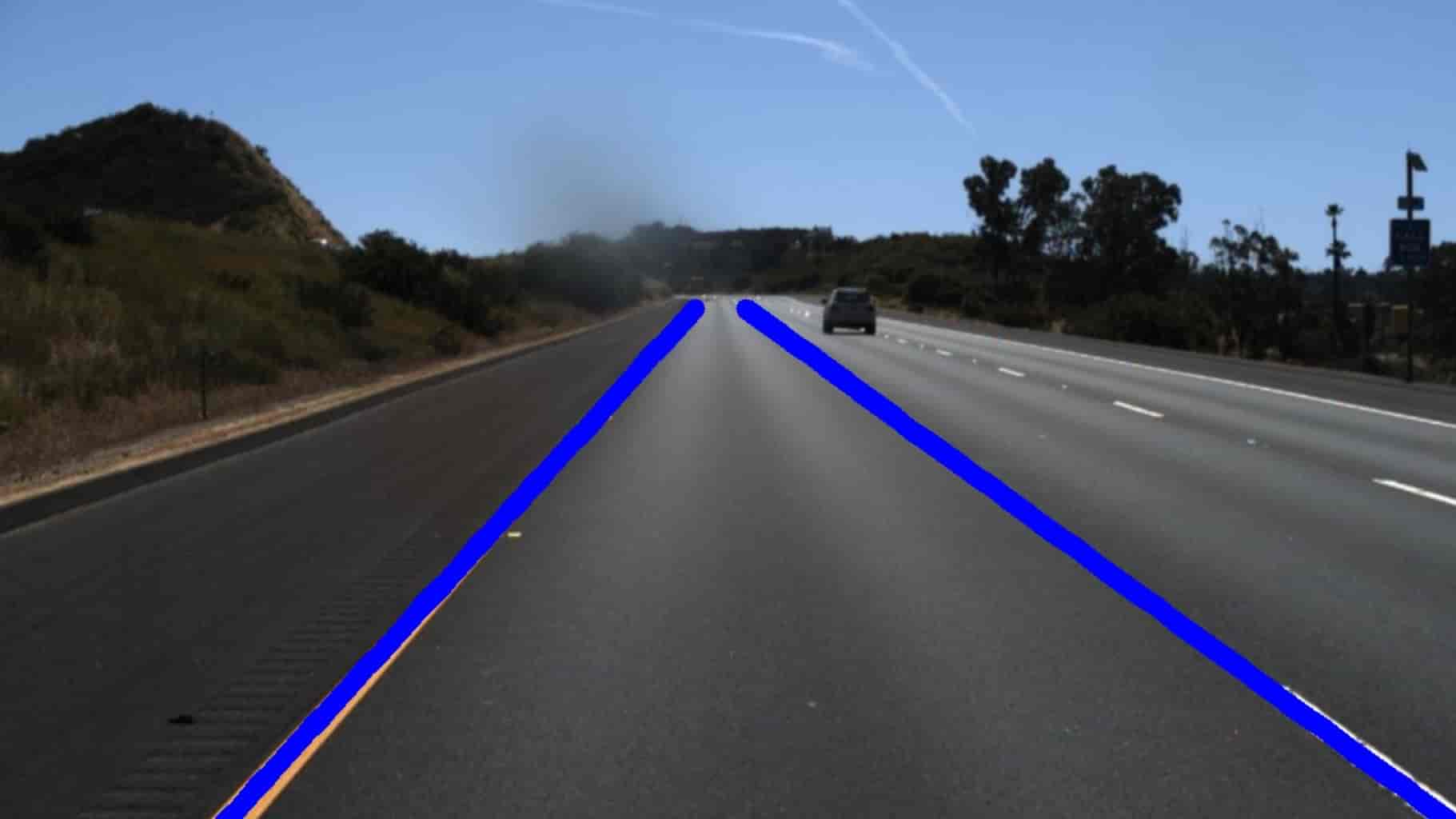}\label{tusimple2pic}
	\vspace{0.4mm}
    
	\subfigure[]{
	    \hspace{-1.3mm}
		\includegraphics[height=18mm, width=39.6mm]{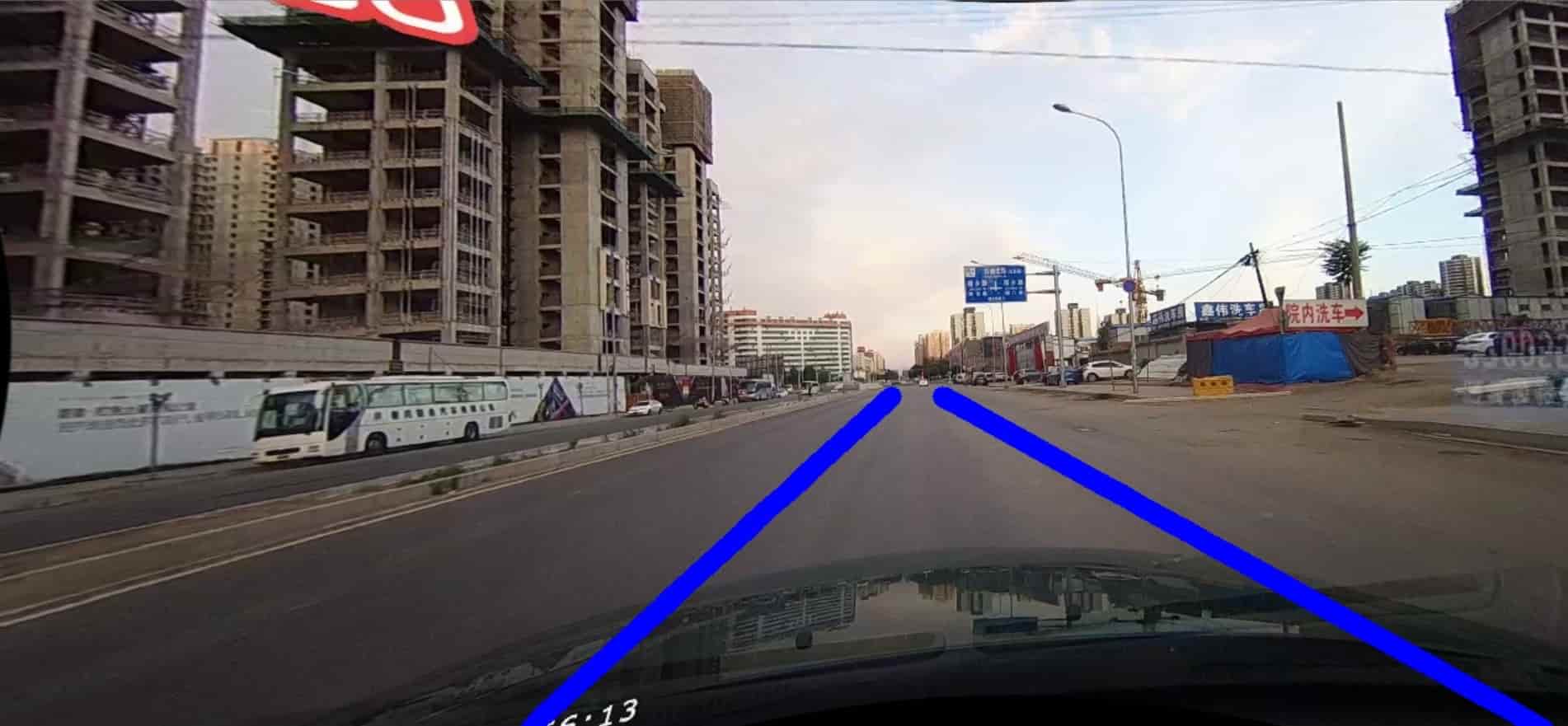}
		\,
	}
	\hspace{-2mm}
	\subfigure[]{
	    \hspace{-2mm}
		\includegraphics[height=18mm, width=32mm]{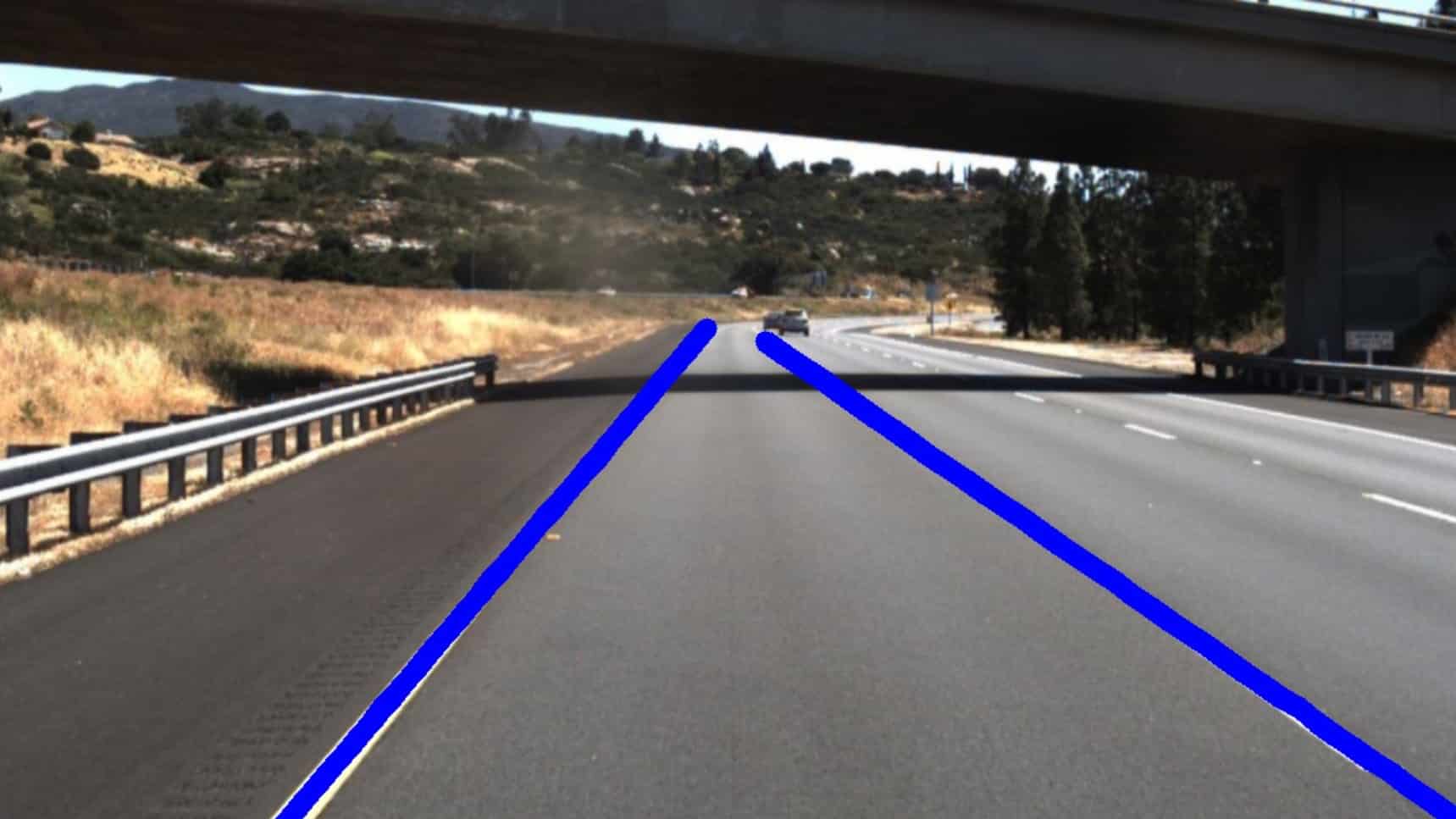}
	}
    \vspace{-1mm}
	\caption{Sample images from the (a) CULane and (b) TuSimple test sets. Ground truth active lane markings are highlighted in blue. }\label{samplepics}\end{figure}

We run experiments on the test sets of two popular and widely used datasets for lane detection, TuSimple~\cite{TuSimple-2019} and CULane~\cite{Pan2018}. Table~\ref{table1} summarizes their details and Fig.~\ref{samplepics} contains sample frames from the datasets. CULane has ground truths labelled on all frames and contains many challenging driving scenarios (\textit{e.g.} congested urban roads and night scenes with poor lighting conditions). TuSimple, on the other hand, is a relatively easy dataset, taken under good or medium weather conditions along highways during the daytime, and only has ground truths labelled on the last frame in each clip of twenty frames. For each frame with ground truths labelled, we manually select the lane markings demarcating the active lane for detection and comparison in our experiments. Some frames in CULane do not have lane markings (\textit{e.g.} when crossing traffic light junctions) and were ignored in our experiments.

\vspace{-0.5mm}
\subsection{Evaluation metrics}
Similar to~\cite{Pan2018,Hou2019}, we calculate intersection over union (IoU) area between the ground truths and predicted lane markings, with line widths set as $16$ and $30$ pixels respectively, for $w=800$ output to identify true positive predicted lane markings. The line widths are adjusted based on model output width for uniform comparison across models. We record results for each IoU threshold between 0.3 and 0.5 (inclusive) at 0.01 intervals. Lane predictions with IoU values above each IoU threshold are marked as true positive (TP) lanes for that threshold level. For our evaluation metric, we employ:
\begin{equation}
    accuracy = \frac{N_\mathit{TP}}{N_{gt}},
\end{equation}
where $N_\mathit{TP}$ is the number of true positive lanes detected at each IoU threshold and $N_{gt}$ is the number of ground truth lanes. For uniform comparison across datasets, we use the same evaluation metric of accuracy across the different datasets.

\begin{table*}
	\small
	\caption{Comparative accuracy results on CULane test set, using CULane-trained models. We include accuracy increase (Acc. Inc.) and the percentage increase in accuracy against our method is in bold.} \label{table2}
	\centering
	\begin{tabular}{c | c c c c c c c |}
		\hline
		IoU threshold
		& SCNN
		& SCNN + RONELD
		& Acc. Inc.
		& ENet-SAD
		& ENet-SAD + RONELD
		& Acc. Inc. \\
		\hline
		0.3
		& 0.812
		& 0.826
		& 0.014 \textbf{(1.7\%)}
		& 0.823
		& 0.832
		& 0.009 \textbf{(1.1\%)} \\
		0.4
		& 0.762
		& 0.789
		& 0.027 \textbf{(3.5\%)}
		& 0.778
		& 0.799
		& 0.021 \textbf{(2.7\%)} \\
		0.5
		& 0.629
		& 0.703
		& 0.074 \textbf{(11.8\%)}
		& 0.655
		& 0.729
		& 0.074 \textbf{(11.3\%)} \\
		\hline
	\end{tabular}
	\vspace{-0.5mm}
\end{table*}
\begin{table*}[ht]
	\small
	\caption{Comparative accuracy results on TuSimple test set, which is a cross-dataset validation using CULane-trained models. We include accuracy increase (Acc. Inc.) and the percentage increase in accuracy against our method is in bold.} \label{table3}
	\centering
	\begin{tabular}{c | c c c c c c c |}
		\hline
		IoU threshold
		& SCNN
		& SCNN + RONELD
		& Acc. Inc.
		& ENet-SAD
		& ENet-SAD + RONELD
		& Acc. Inc. \\
		\hline
		0.3
		& 0.625
		& 0.869
		& 0.244 \textbf{(39.0\%)}
		& 0.608
		& 0.825
		& 0.217 \textbf{(35.7\%)} \\
		0.4
		& 0.470
		& 0.796
		& 0.326 \textbf{(69.4\%)}
		& 0.502
		& 0.753
		& 0.251 \textbf{(50.0\%)} \\
		0.5
		& 0.238
		& 0.549
		& 0.311 \textbf{(130.7\%)}
		& 0.341
		& 0.530
		& 0.189 \textbf{(55.4\%)} \\
		\hline
	\end{tabular}
	\vspace{-0.5mm}
\end{table*}

\subsection{Implementation details}
We exploit two state-of-the-art methods, namely SCNN~\cite{Pan2018} and ENet-SAD~\cite{Hou2019}, for comparison with our RONELD method. The models are pre-trained with the CULane train set and we deliberately do not include any images from the TuSimple dataset in the train set for cross-dataset validation. We use the CULane-trained SCNN and ENet-SAD models to generate probability map outputs on the CULane and TuSimple test set images. From these probability maps, we use the method outlined in~\cite{Pan2018} and~\cite{Hou2019} to generate lane markings for the SCNN and ENet-SAD model for comparison. The method searches every twenty rows in the probability map, selects the highest confidence point as a lane point, and connects them using cubic splines to obtain lane marking predictions. To obtain lane marking predictions for SCNN + RONELD and ENet-SAD + RONELD, we run RONELD on the same probability map outputs obtained from the CULane-trained SCNN and ENet-SAD models respectively on the CULane and TuSimple test set. We compare the lane marking predictions with the ground truths and compute the corresponding accuracy results at the different IoU thresholds for the various methods.

\begin{figure}
    \centering
    \subfigure[Accuracy results on CULane test set using CULane-trained models.]{
        \hspace{-4.5mm}
        \includegraphics[width=0.90\columnwidth]{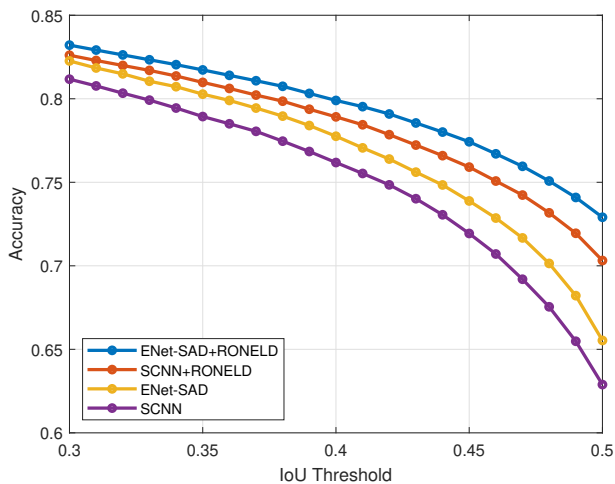}\label{culane_graph}
    }
    \vspace{4mm}
    \subfigure[Accuracy results on TuSimple test set, which is a cross-dataset validation using CULane-trained models.]{
        \includegraphics[width=0.89\columnwidth]{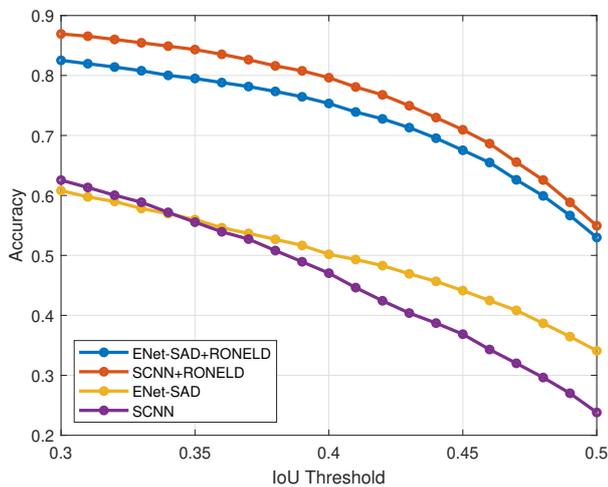}\label{tusimple_graph}
    }
    \vspace{2mm}
    \caption{Comparative accuracy results for \protect\subref{culane_graph} CULane and \protect\subref{tusimple_graph} TuSimple test sets using CULane-trained state-of-the-art deep learning models, with and without RONELD, between 0.3 to 0.5 IoU thresholds at 0.01 IoU threshold intervals.}\label{graphFig}
\end{figure}

\subsection{Results}
\begin{figure*}
	\subfigure[SCNN]
	{
		\includegraphics[width=42mm]{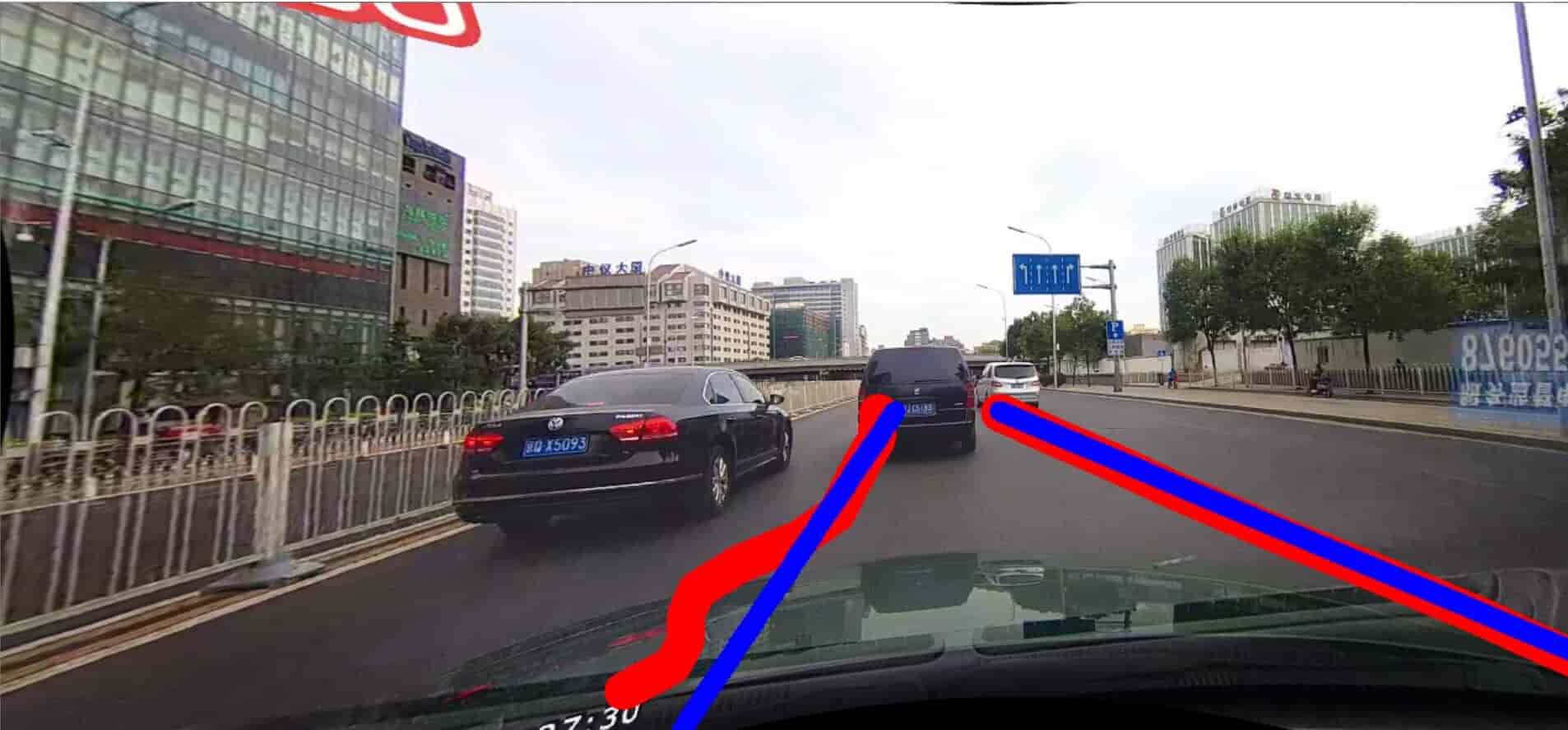}\label{fig3a}
	}
	\vspace{-0.7mm}
	\subfigure[SCNN + RONELD]
	{
		\includegraphics[width=42mm]{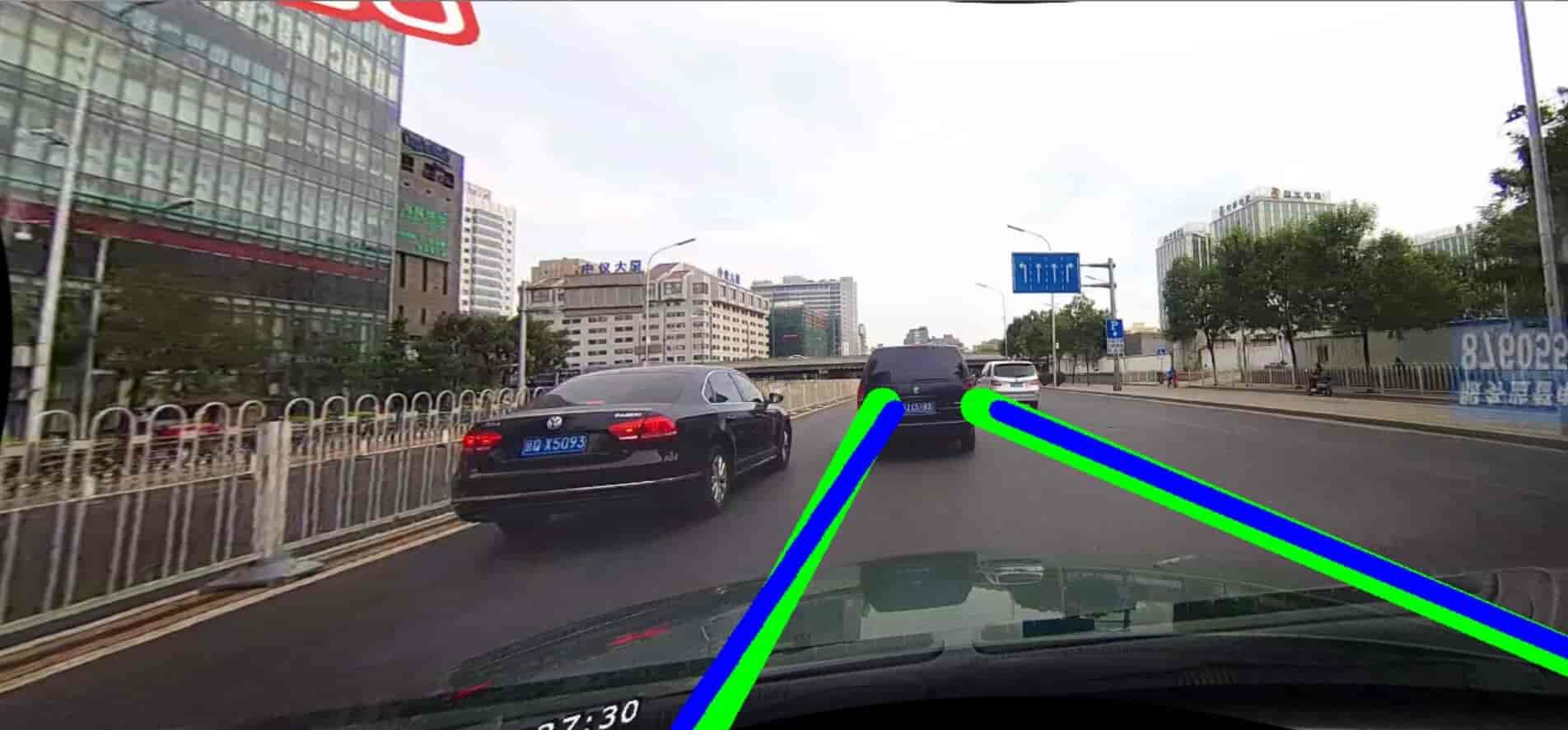}\label{fig3b}
	}
	\vspace{-0.7mm}
	\subfigure[ENet-SAD]
	{
		\includegraphics[width=42mm]{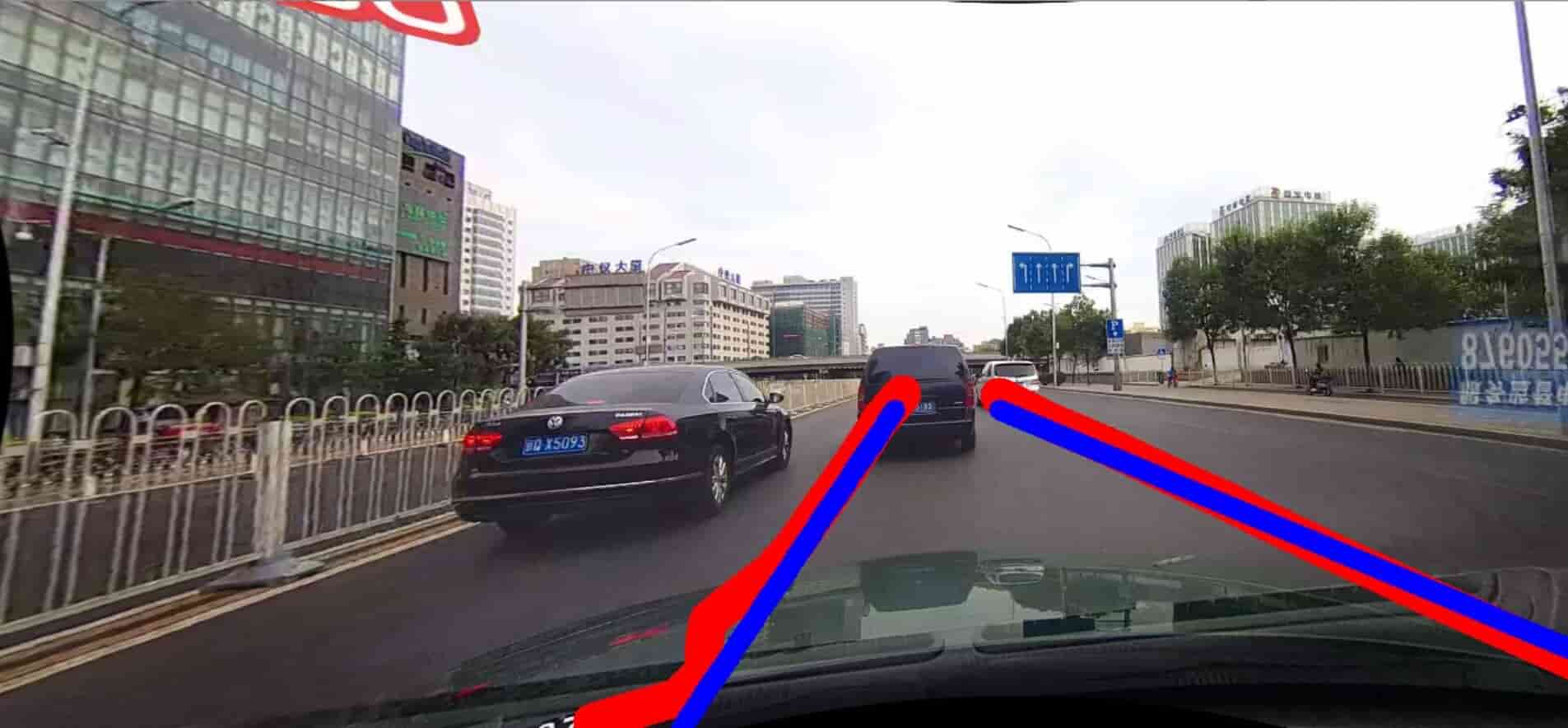}\label{fig3c}
	}
	\vspace{-0.7mm}
	\subfigure[ENet-SAD + RONELD]
	{
		\includegraphics[width=42mm]{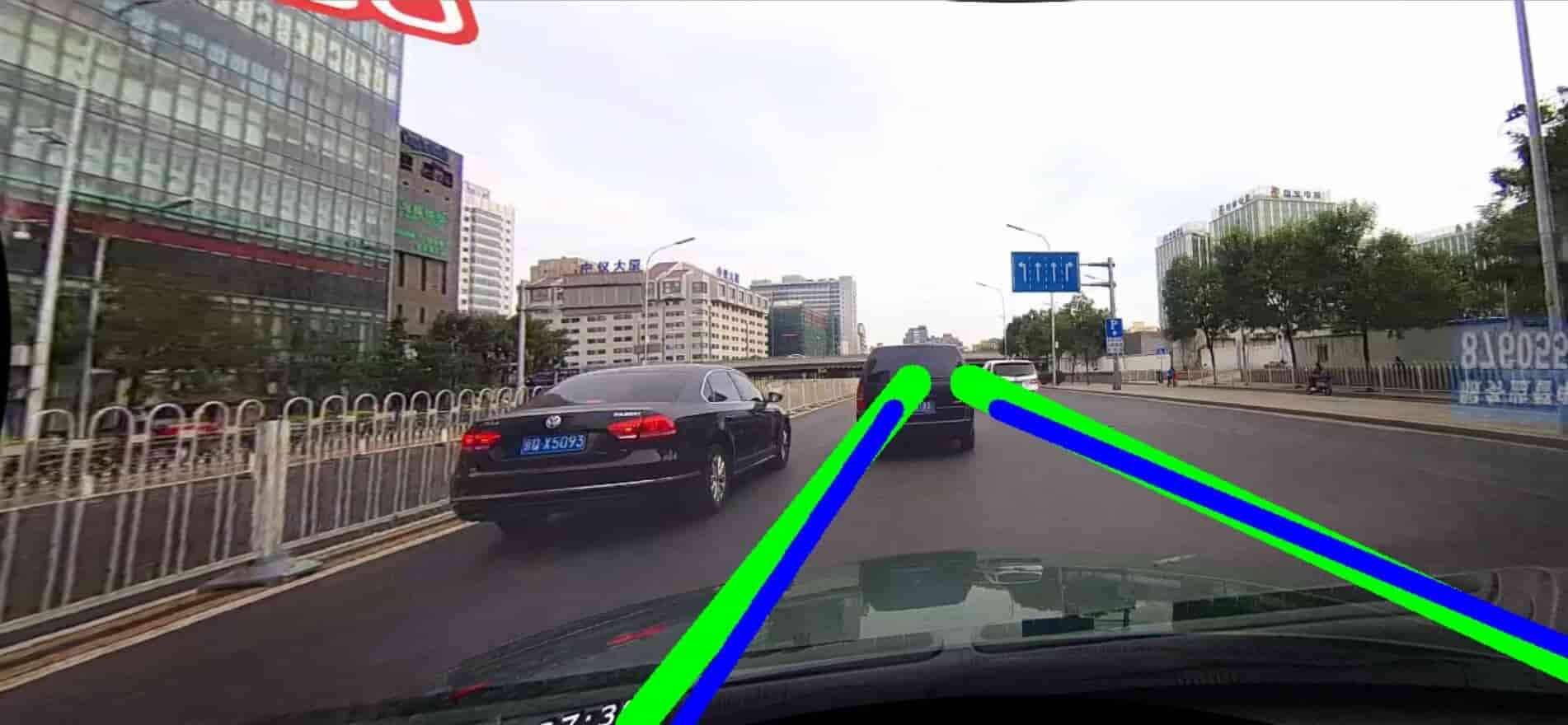}\label{fig3d}
	}
	\vspace{-0.7mm}
	\subfigure[SCNN]
	{
		\includegraphics[width=42mm]{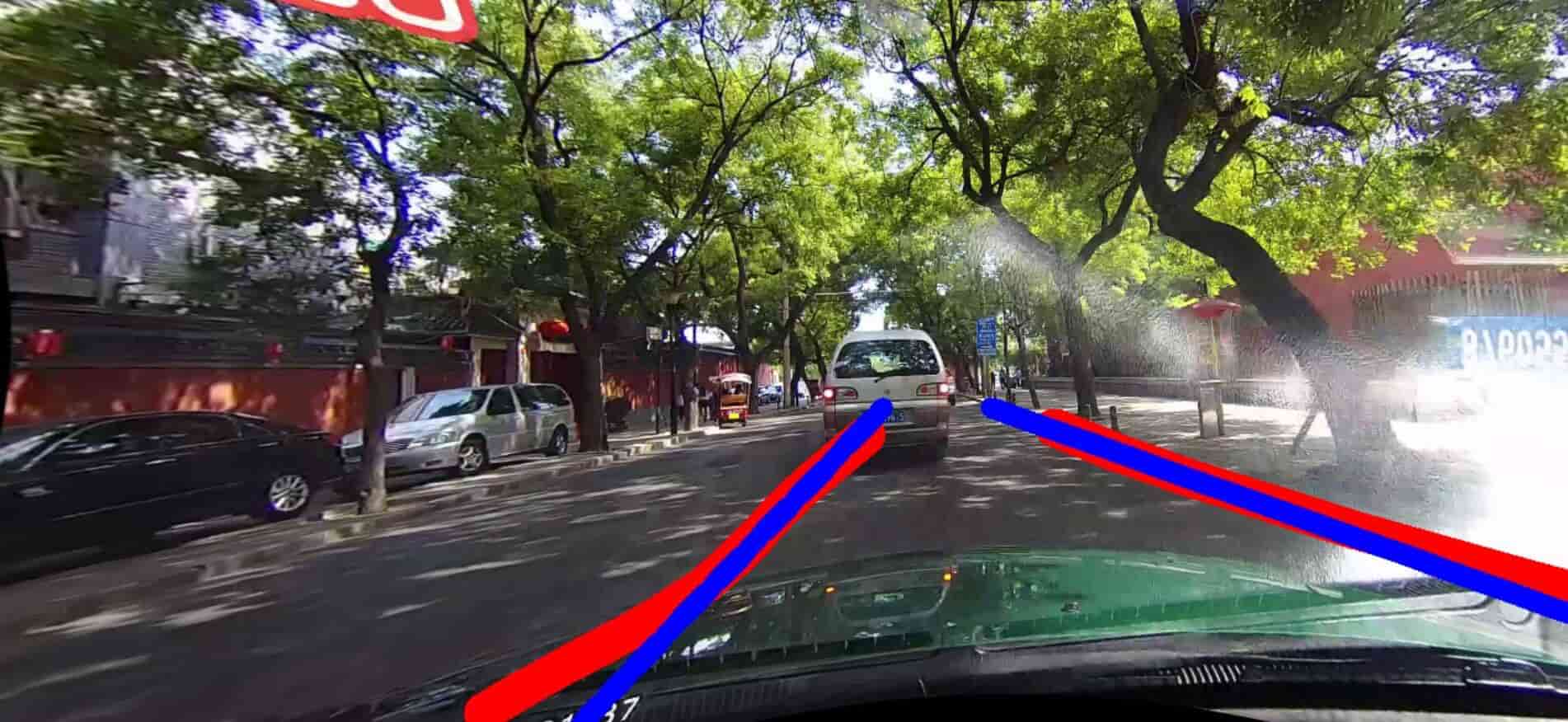}\label{fig3e}
	}
	\subfigure[SCNN + RONELD]
	{
		\includegraphics[width=42mm]{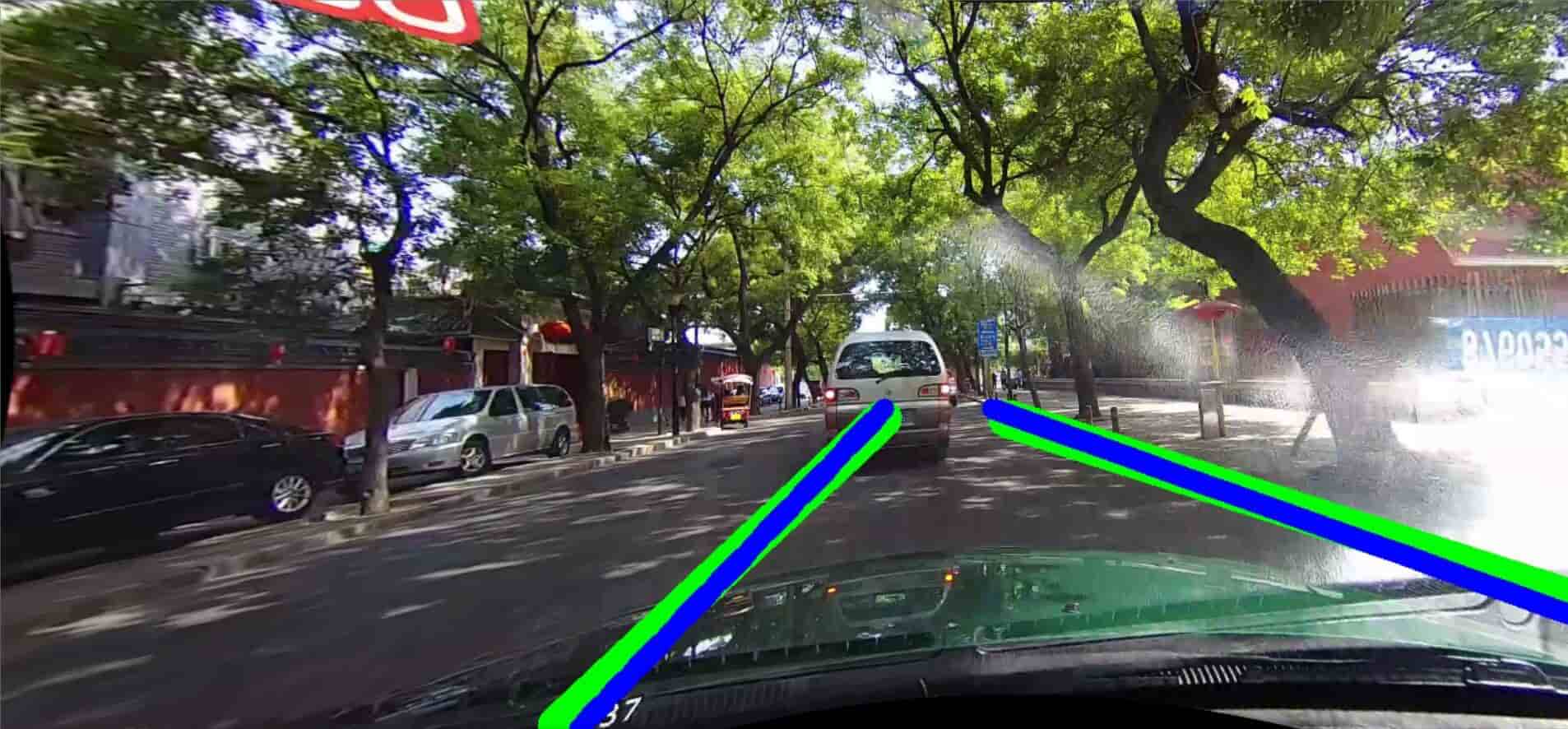}\label{fig3f}
	}
	\subfigure[ENet-SAD]
	{
		\includegraphics[width=42mm]{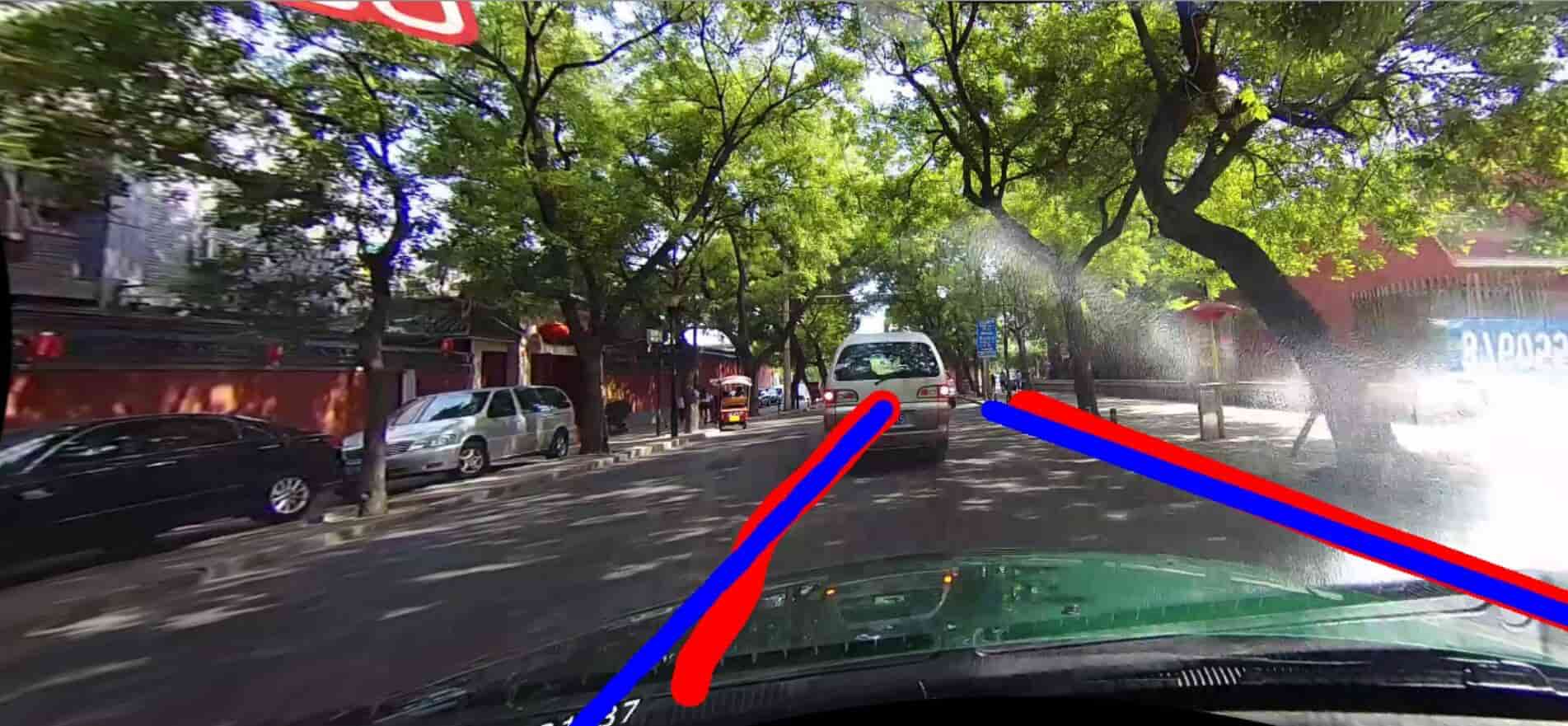}\label{fig3g}
	}
	\subfigure[ENet-SAD + RONELD]
	{
		\includegraphics[width=42mm]{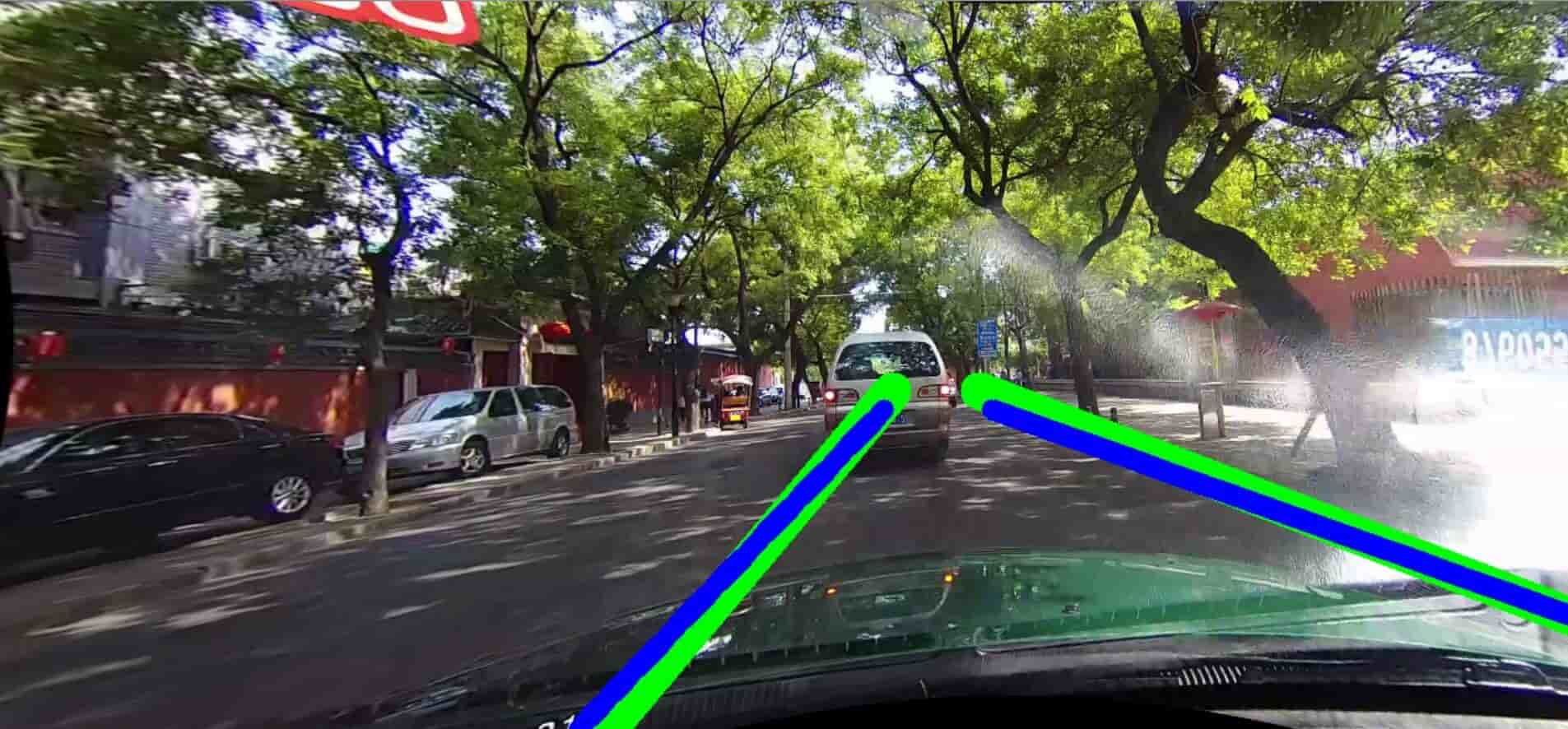}\label{fig3h}
	}
\subfigure[SCNN]
{
	\includegraphics[width=42mm]{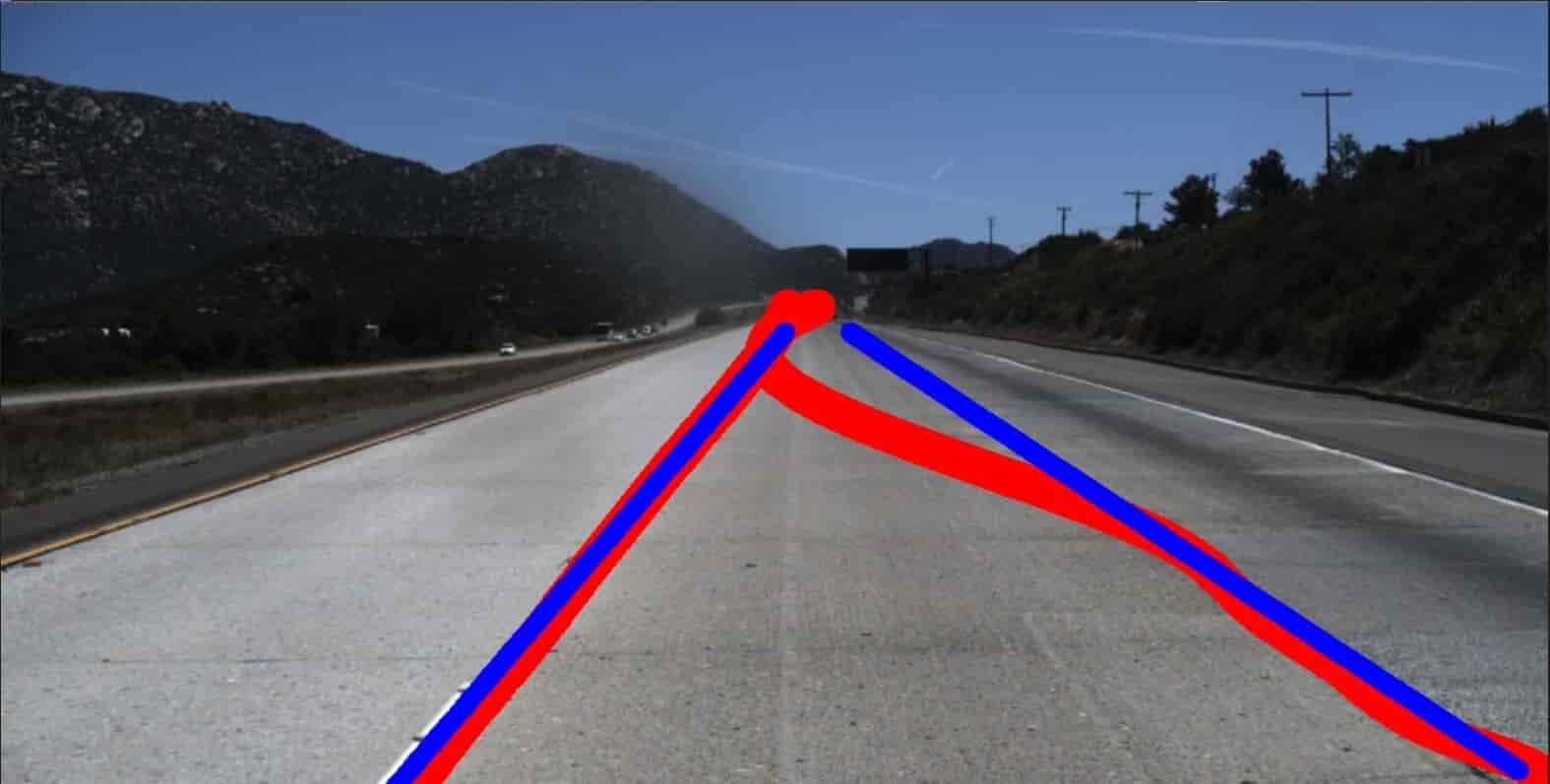}\label{fig3i}
}
\subfigure[SCNN + RONELD]
{
	\includegraphics[width=42mm]{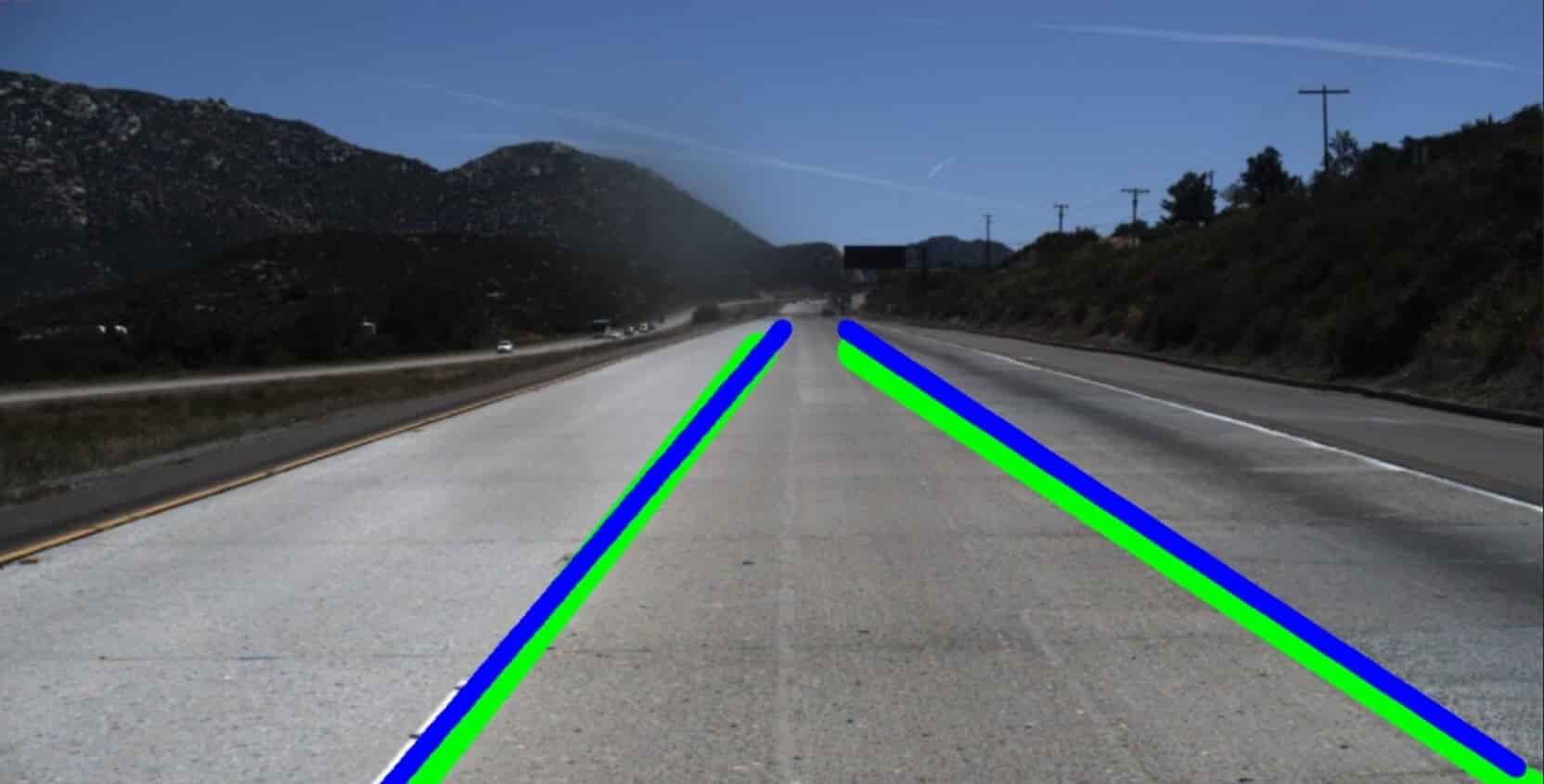}\label{fig3j}
}
\subfigure[ENet-SAD]
{
	\includegraphics[width=42mm]{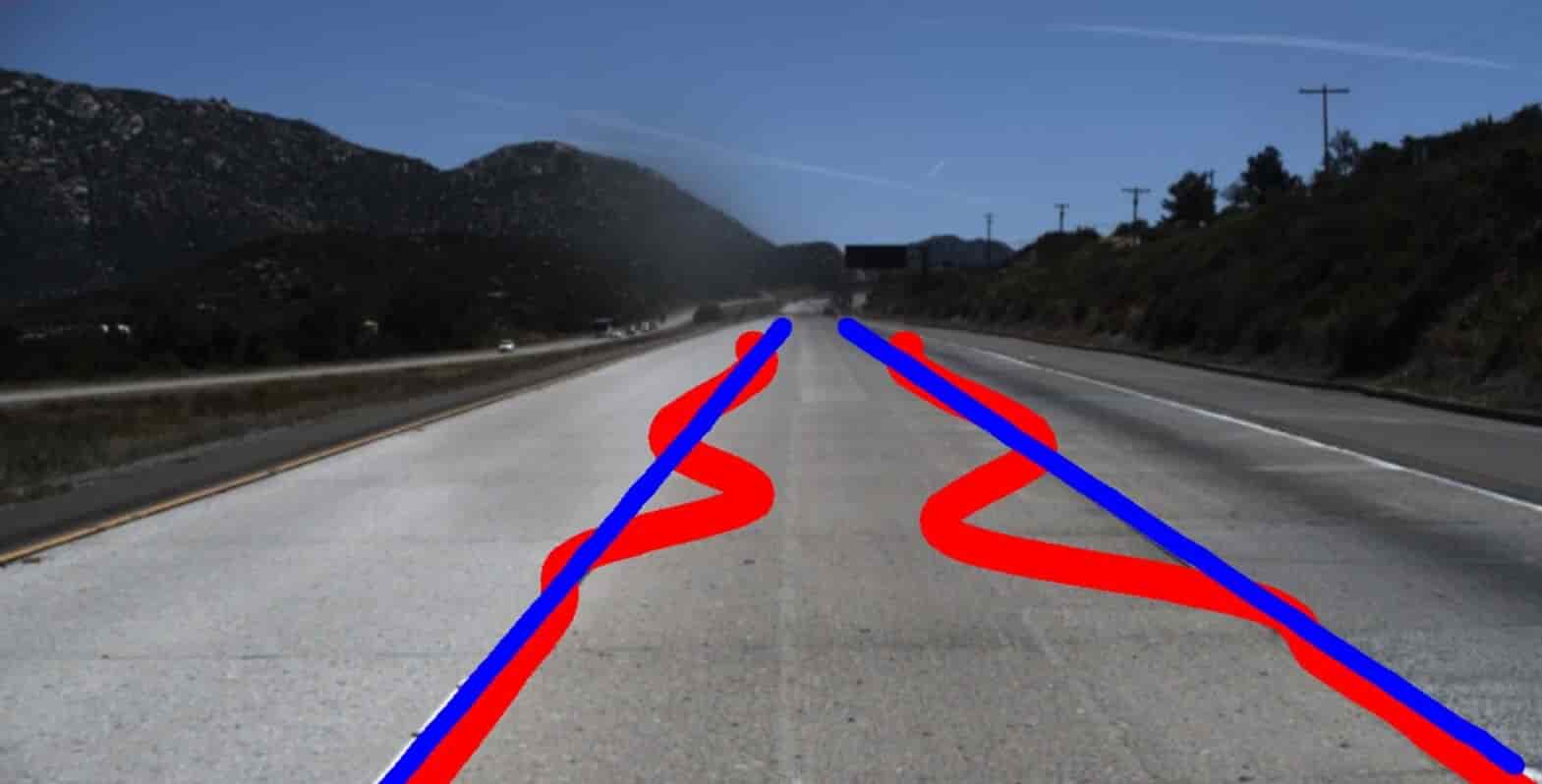}\label{fig3k}
}
\subfigure[ENet-SAD + RONELD]
{
	\includegraphics[width=42mm]{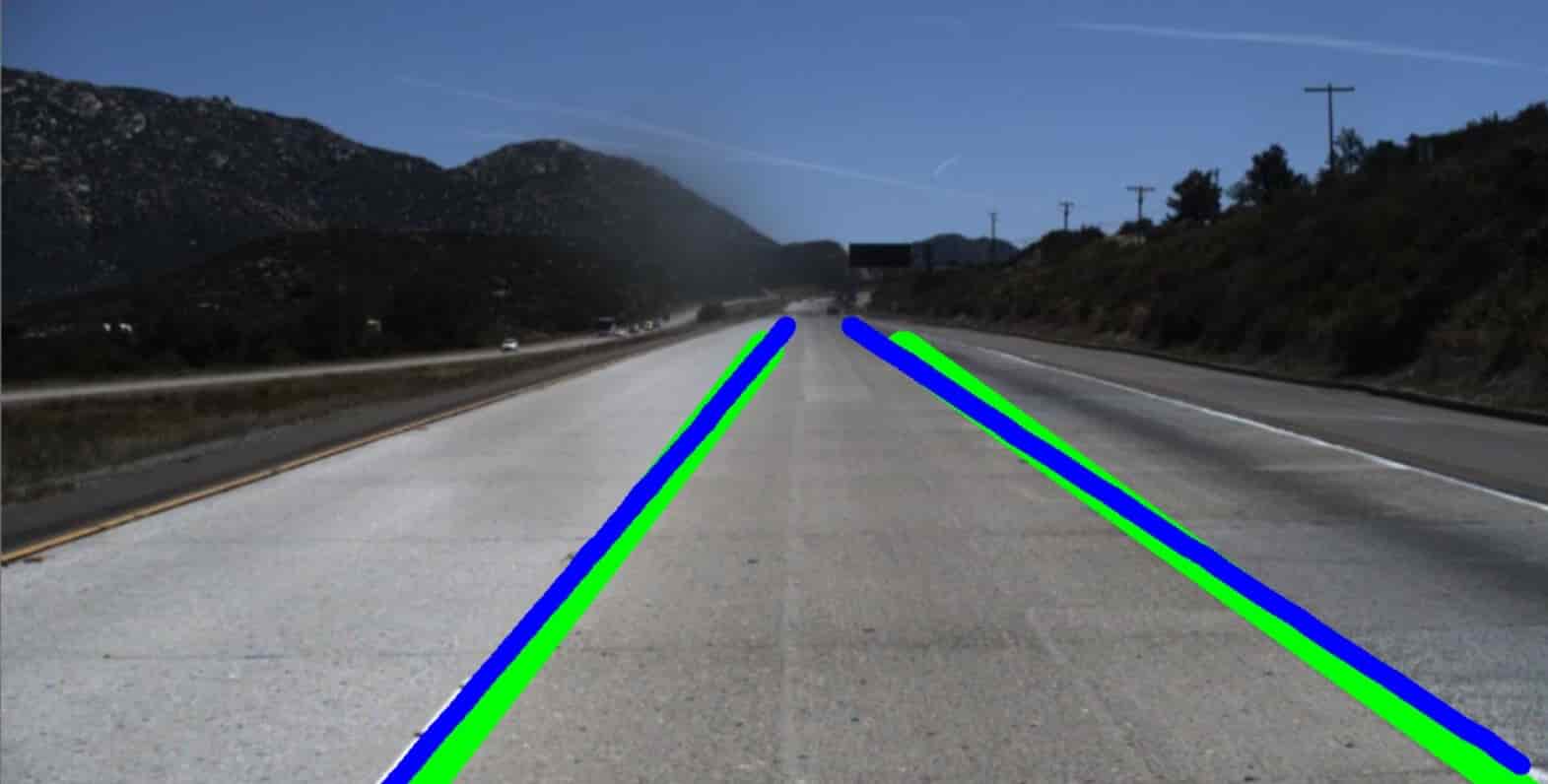}\label{fig3l}
}
	\caption{Lane detection results for SCNN, SCNN + RONELD, ENet-SAD and ENet-SAD + RONELD are shown.  (a), (b), (c), (d), (e), (f), (g), (h) are from the CULane dataset, and (i), (j), (k), (l) are from the TuSimple dataset. Ground truths are highlighted in blue, lane detection results from the state-of-the-art algorithms are in red and corresponding results with RONELD are highlighted in green.}\label{fig3}
	\end{figure*}

Tables~\ref{table2}-\ref{table3} and Fig.~\ref{graphFig} summarize the accuracy performance of our methods, \textit{i.e.} SCNN + RONELD and ENet-SAD + RONELD, against SCNN and ENet-SAD respectively, on the CULane and TuSimple test sets. Tables~\ref{table2}-\ref{table3} show accuracy results and percentage increase in accuracy for the 0.3, 0.4 and 0.5 IoU thresholds, which correspond to loose, medium and strict evaluations respectively. Fig.~\ref{graphFig} shows accuracy results for 0.3 to 0.5 IoU thresholds (inclusive) at 0.01 intervals. Some comparative imaging results are shown in Fig.~\ref{fig3}, and our discussion of the results are as follows.


\textbf{CULane results.} It is observed that improvements using our RONELD method are not striking on this particular dataset, except the $11.8\%$ and $11.3\%$ increase in accuracy on the strictest $0.5$ IoU threshold for SCNN and ENet-SAD respectively. The high degree of similarity between the CULane test and train sets, which are from the same city, explains the general good performance of both deep learning models, with and without RONELD, on the CULane test set. This results in less room for RONELD to improve on the lane detection outputs of the existing models. Furthermore, it is observed that errors in the lane detection outputs are mostly due to incorrectly identified lanes in the deep learning model probability map output, similar to Fig.~\ref{fig1c}, which is difficult for RONELD to address while remaining robust to changes in the lane markings from the  probability map output. Despite this, our experiments show that adding RONELD leads to a positive increase in accuracy performance of the SCNN and ENet-SAD models on all IoU thresholds tested.

\textbf{TuSimple results.} We see a more significant accuracy improvement on this dataset after using our RONELD method. It is apparent that the state-of-the-art algorithms do not work well on the unseen TuSimple test set in cross-dataset validation tests, particularly at higher IoU thresholds. By applying RONELD, we are able to achieve compelling results, with a $35.7\%$ to $69.4\%$ increase in accuracy on the looser $0.3$ and $0.4$ IoU thresholds. More significantly, the increase in accuracy on the strictest $0.5$ IoU threshold is above 50\% for ENet-SAD and two-fold for SCNN. Furthermore, a common issue with the SCNN and ENet-SAD model probability map output on the unseen TuSimple dataset appears to be undetected lanes in intermediate frames and a high degree of noise causing distorted lanes. These arise due to the TuSimple test set differing in some significant ways from the CULane test set (\textit{e.g.} road surface conditions, types of lane markings). To address these, RONELD uses linear regression for detected straight lanes and outlier removal to reduce noise and through tracking of preceding frames, RONELD can map lanes from previous frame to the current frame to address the problem of undetected lanes in some intermediate frames. As a result, RONELD has a more significant increase in accuracy performance on the TuSimple dataset, with more stable active lanes that are less susceptible to noise. This makes them more suitable for autonomous driving applications as compared to the lanes detected from the deep learning models as shown in Fig.~\ref{fig3}(i), (j), (k), (l).

\textbf{Discussion.} It is observed that adding RONELD to the deep learning models improves accuracy performance at all IoU thresholds measured. Interestingly, on the looser $0.3$ and $0.4$ IoU thresholds, SCNN + RONELD achieves a higher accuracy performance on the unseen TuSimple test set \textit{vis a vis} the CULane test set, while ENet-SAD + RONELD achieves comparable performance on the 0.3 IoU threshold for both test sets, despite both SCNN and ENet-SAD models being trained on the CULane train set. This better performance is explained by TuSimple being a relatively simple dataset compared to CULane as well as each clip in the TuSimple test set containing nineteen preceding frames for all labelled images, with only the last frame in each twenty-frame clip containing ground truth lane markings. This provides RONELD with preceding frames to process before being compared with ground truth frames and highlights the ability of RONELD to effectively utilize lane information in preceding frames.

\textbf{Ablation study.} To investigate our RONELD method and verify its effectiveness, we completed an ablation study to understand the effect of the preceding frame tracking (PFT) step in section~\ref{sec:methodology}. To test this, we run RONELD with and without PFT by controlling information passing between different RONELD method calls. Our results are shown in Table~\ref{PFTtable}. It is observed that the increase due to PFT is significantly larger at the 0.3 IoU threshold compared to the 0.5 IoU threshold, which is expected as lanes in preceding frames are less likely to have the high accuracy needed to meet the higher threshold, but provide a good estimate of the current lane position. The increase in accuracy is also observed to be more significant on the TuSimple test set \textit{vis a vis} the CULane test set. This is also in line with expectations due to the greater number of undetected lanes on the TuSimple dataset as it is a cross-dataset validation using the CULane-trained models.

\begin{table}
  \begin{center}
      \caption{Accuracy results of SCNN + RONELD with and without preceding frame tracking (PFT) at 0.3 and 0.5 IoU thresholds.}
      \label{PFTtable}
      \begin{tabular}{c|c c c c}
        \hline
        \multirow{2}{*}{IoU Threshold} &
          \multicolumn{2}{c}{CULane} &
          \multicolumn{2}{c}{TuSimple} \\
          \cline{2-5}
          & {w/o PFT} & {w/ PFT} & {w/o PFT} & {w/ PFT}\\
          \hline
        0.3 & 0.818 & 0.826 & 0.773 & 0.869 \\
        0.5 & 0.702 & 0.703 & 0.502 & 0.549 \\
        \hline
    \end{tabular}
  \end{center}
\end{table}
\begin{table}[t]
  \begin{center}
    \caption{Average runtime of SCNN + RONELD and ENet-SAD + RONELD on the CULane and TuSimple test sets (in milliseconds) using a Python 3 + Numba~\cite{Lam2015} implementation.}
    \label{timingtable}
    \begin{tabular}{c | c  c} 
        \hline
		Dataset
		& SCNN + RONELD
		& ENet-SAD + RONELD  \\
		\hline
		CULane
		& 5.68
		& 6.29\\
		TuSimple
		& 2.80
		& 3.55 \\
		\hline
		Mean
		& 4.24
		& 4.92 \\
		\hline
    \end{tabular}
  \end{center}
\end{table}

\textbf{Runtime.} To verify RONELD's fast runtime for real-time use on autonomous vehicles, we measured the average runtime of RONELD on a single Intel Core i9-9900K CPU by taking the mean time needed for RONELD to process the probability maps from the deep learning models as inputs and return lane markings across all images in the test sets, including test set images without labelled ground truths. Using a Python 3 + Numba~\cite{Lam2015} implementation, the average runtimes are recorded in Table~\ref{timingtable}. It is observed that the average runtime on the TuSimple test set is noticeably lower than that of the CULane test set. This is due to the lower number of detected lane points per image on the TuSimple test set by the CULane-trained models, as TuSimple is an unseen dataset for the models which are therefore less able to detect lane markings on the TuSimple test set. This is reflected by the significantly weaker performance of the deep learning models on the TuSimple test set, both with and without RONELD. Less points detected per lane image requires less processing by RONELD, resulting in shorter runtimes on the TuSimple test set. The difference in average runtimes on the ENet-SAD and SCNN models can be explained by the larger size of the ENet-SAD probability map. In general, it can be seen that RONELD is a low computational time method, suitable to be paired with deep learning models for real-time use on autonomous vehicles, with overall mean average runtimes of less than 5ms on both deep learning models tested.

\section{CONCLUSION}
\label{sec:conclusion}

We have presented a robust neural network output enhancement for active lane detection (RONELD) method which achieves compelling results in cross-dataset validation tests and shows high potential for use in real-time autonomous driving applications. Using RONELD, we identify, track and optimize active lane detection on probability maps from deep learning based lane detection algorithms. We have demonstrated on two state-of-the-art algorithms, tagged as SCNN + RONELD and ENet-SAD + RONELD, in our experiments on the CULane and TuSimple datasets. Results over the two datasets indicated that by applying RONELD, accuracy increases by up to $69.4\%$ on the looser $0.3$ and $0.4$ IoU thresholds, and increases up to two-fold on the strictest 0.5 IoU threshold against both SCNN and ENet-SAD algorithms.

\bibliographystyle{IEEEtran}
\bibliography{refs}

\end{document}